%% file: main.tex
\documentclass{article}
\usepackage[utf8]{inputenc}
\usepackage{biblatex}
\usepackage{braket,amsfonts}
\usepackage{amsmath}
\usepackage{amsthm}
\usepackage{amssymb}
\usepackage[LGR,T1]{fontenc}

\newtheorem{lemma}{Lemma}[section]
\newtheorem{theorem}{Theorem}[section]
\newtheorem{assumption}{Assumption}[section]
\newtheorem{proposition}{Proposition}[section]
\newtheorem{remark}{Remark}[section]
\newtheorem{definition}{Definition}[section]
\usepackage{array}
\usepackage{parskip}
\usepackage[caption=false]{subfig}
\usepackage{pgfplots}

\usepackage{mathtools, nccmath}
\usepackage{graphicx,epstopdf}
\usepackage{enumerate}
\usepackage{amsmath}

\usepackage{amsopn}

\allowdisplaybreaks
\usepackage[hidelinks]{hyperref}
\newcommand*{\email}[1]{\href{mailto:#1}{\nolinkurl{#1}} } 
\usepackage{cleveref}
\usepackage{xspace}
\usepackage{makecell}
\usepackage{bold-extra}
\usepackage[most]{tcolorbox}

\colorlet{texcscolor}{blue!50!black}
\colorlet{texemcolor}{red!70!black}
\colorlet{texpreamble}{red!70!black}
\colorlet{codebackground}{black!25!white!25}

\newcommand{\mix}{\text{mix}}
\newcommand{\tv}{\text{TV}}

\lstdefinestyle{siamlatex}{%
  style=tcblatex,
  texcsstyle=*\color{texcscolor},
  texcsstyle=[2]\color{texemcolor},
  keywordstyle=[2]\color{texemcolor},
  moretexcs={cref,Cref,maketitle,mathcal,text,headers,email,url},
}

\tcbset{%
  colframe=black!75!white!75,
  coltitle=white,
  colback=codebackground, 
  colbacklower=white, 
  fonttitle=\bfseries,
  arc=0pt,outer arc=0pt,
  top=1pt,bottom=1pt,left=1mm,right=1mm,middle=1mm,boxsep=1mm,
  leftrule=0.3mm,rightrule=0.3mm,toprule=0.3mm,bottomrule=0.3mm,
  listing options={style=siamlatex}
}

\newtcblisting[use counter=example]{example}[2][]{%
  title={Example~\thetcbcounter: #2},#1}

\newtcbinputlisting[use counter=example]{\examplefile}[3][]{%
  title={Example~\thetcbcounter: #2},listing file={#3},#1}

\DeclareTotalTCBox{\code}{ v O{} }
{ 
  fontupper=\ttfamily\color{black},
  nobeforeafter,
  tcbox raise base,
  colback=codebackground,colframe=white,
  top=0pt,bottom=0pt,left=0mm,right=0mm,
  leftrule=0pt,rightrule=0pt,toprule=0mm,bottomrule=0mm,
  boxsep=0.5mm,
  #2}{#1}

\patchcmd\newpage{\vfil}{}{}{}
\graphicspath{{Images/}}

\usepackage{import}
\usepackage{mathtools}

\input{math_commands}

\usepackage{algpseudocode}
\usepackage{algorithm}

\title{Finite-Time Analysis of Asynchronous Q-learning under Diminishing Step-Size from Control-Theoretic View}

\author{Han-Dong Lim\thanks{Department of Electrical and Engineering, Korea Advanced Institute of Science and Technology (KAIST),
Daejeon, 34141, South Korea (\email{limaries30@kaist.ac.kr},\email{donghwan@kaist.ac.kr})} \and Donghwan Lee\footnotemark[1]
\thanks{This work was supported by Institute of Information communications Technology Planning Evaluation (IITP) grant  funded by the Korea government (MSIT)(No.2022-0-00469)}
} 

\begin{document}

\maketitle

\begin{tcbverbatimwrite}{tmp_\jobname_abstract.tex}
\begin{abstract}
Q-learning has long been one of the most popular reinforcement learning algorithms, and theoretical analysis of Q-learning has been an active research topic for decades. Although researches on asymptotic convergence analysis of Q-learning have a long tradition, non-asymptotic convergence has only recently come under active study. The main goal of this paper is to investigate new finite-time analysis of asynchronous Q-learning under Markovian observation models via a control system viewpoint. In particular, we introduce a discrete-time time-varying switching system model of Q-learning with diminishing step-sizes for our analysis, which significantly improves recent development of the switching system analysis with constant step-sizes, and leads to \(\mathcal{O}\left( \sqrt{\frac{\log k}{k}} \right)\) convergence rate that is comparable to or better than most of the state of the art results in the literature. In the mean while, a technique using the similarly transformation is newly applied to avoid the difficulty in the analysis posed by diminishing step-sizes. The proposed analysis brings in additional insights, covers different scenarios, and provides new simplified templates for analysis to deepen our understanding on Q-learning via its unique connection to discrete-time switching systems.
\end{abstract}

\end{tcbverbatimwrite}
\input{tmp_\jobname_abstract.tex}
\section{Introduction}
    \import{./introduction}{introduction.tex}
\section{Preliminaries}\label{sec:prelim}
    \subsection{Notations}
        \import{./preliminaries}{notation.tex}

    \subsection{Markov decision process}
    \import{./preliminaries}{markov_decision_process}

    \subsection{Switching system}

\import{./preliminaries}{switching_system.tex}
    \subsection{Linear time-varying system}
        \import{./preliminaries}{ltv.tex}
    \subsection{Matrix notations}

\import{./preliminaries}{qlearning_notations.tex}
    \subsection{Revisit Q-learning}\label{sec:revisit-Q-learning}

\import{./preliminaries}{revisit_qlearning.tex}
    \subsection{Linear stochastic approximation}\label{sec:lsa}
        \import{./preliminaries}{linear_sa.tex}
 \section{Control system approach}\label{sec:csa}

\import{./preliminaries}{control_system_approach.tex}

\section{Analysis of lower comparison system}\label{sec:ls}
    \import{./lower_system}{header.tex}
    \subsection{Construction of lower comparison system}
        \import{./lower_system}{construction.tex}

    \subsection{Analysis of lower comparison system}
    \import{./lower_system}{lower_system.tex}

\section{Upper comparison system}\label{sec:us}
    \import{./}{upper_system.tex}

\section{Original system}\label{sec:os}
    \import{./}{original_system.tex}
\section{Finite time analysis under Markovian noise}\label{sec:fta_markov}
    \import{markovian}{markovian_copy.tex}

\section{Conclusion}\label{sec:conclusion}
    \import{./conclusion}{conclusion.tex}

\printbibliography
\section{Appendix}
    \import{./appendix}{appendix.tex}\import{./appendix}{common_lyapunov.tex}
\end{document}

%% file: math_commands.tex
\DeclareMathOperator*{\argmax}{arg\,max}

%% file: introduction/introduction.tex
\par 

Recently, reinforcement learning has shown remarkable performance in complex and challenging domains. In the seminal work~\Cite{mnih2015human}, the so-called deep Q-learning has achieved human level performance in challenging tasks such as the Atari video games using deep learning based approaches. Since then, there have been major improvements~\Cite{kapturowski2018recurrent,schrittwieser2020mastering,badia2020never,badia2020agent57} to solve the Atari benchmark problems as well. On the other hand, significant progresses have been made for other various fields such as recommendation system~\Cite{afsar2021reinforcement}, robotics~\Cite{kober2013reinforcement}, and portfolio management~\Cite{jiang2017cryptocurrency} to name just a few.
\par
  Developed by Watkins in~\Cite{watkins1992q}, Q-learning is one of the most widely known reinforcement learning algorithms. Theoretical analysis of Q-learning has been an active research topic for decades. In particular, asymptotic convergence of Q-learning has been successfully established through a series of works~\Cite{tsitsiklis1994asynchronous,jaakkola1993convergence,borkar2000ode,lee2020unified} based on stochastic approximation theory~\Cite{borkar2000ode}. 
  Although the asymptotic analysis ensures the essential property, the convergence to a solution, the natural question remains unanswered: does the algorithm make a consistent and quantifiable progress toward an optimal solution?
  To understand this non-asymptotic behavior, meaningful advances have been made recently~\Cite{szepesvari1997asymptotic,even2003learning,lee2021discrete,li2020sample,chen2021lyapunov,qu2020finite,li2021tightening} as well. Among the promising results, our main concern is the control-theoretic framework~\Cite{lee2021discrete}, which provides a unique switching system~\Cite{liberzon2003switching} perspective of Q-learning. In particular, asynchronous Q-learning with a constant step-size has been naturally formulated as a stochastic discrete-time affine switching system, which has allowed us to transform the convergence analysis into a stability analysis of the switching system model from control theoretic viewpoints. This analysis provides new insights and analysis framework to deepen our understanding on Q-learning via its unique connection to discrete-time switching systems with remarkably simplified proofs. Nevertheless, this work leaved key questions unanswered, which are summarized as follows: 
  \begin{enumerate}
      \item The study in~\Cite{lee2021discrete} provides a finite-time analysis for the averaged iterate of Q-learning rather than the final-time iterate. However, the averaged iterate requires additional steps, and it can slow down the convergence in practice.
      
      \item It only considers a constant step-size, which results in biases in the final solution. On the other hand, a diminishing step-size is more widely used in theoretical analysis to guarantee convergence to a solution without the biases. 

      \item The analysis in~\Cite{lee2021discrete} adopts i.i.d. observation models. Although this assumption simplifies the overall analysis, it is too idealistic to reflect real world scenarios. 
  \end{enumerate}
  
 Motivated from the above discussions, the main goal of this paper is to revisit the non-asymptotic analysis of asynchronous Q-learning~\Cite{lee2021discrete} from discrete-time switching system viewpoints. Contrary to~\Cite{lee2021discrete}, we consider diminishing step-sizes, Markovian observation models, and the final-time convergence. Our main contributions are summarized below. 
\paragraph{Contributions}
\begin{enumerate}[1)]
 \item A new finite-time analysis is proposed for asynchronous Q-learning under Markovian observation models and diminishing step-sizes, which answers the key questions that were left unsolved in~\Cite{lee2021discrete}. The new analysis offers \( \mathcal{O}\left( \sqrt{\frac{\log k}{k}}\right) \) convergence rate in terms of number of iterations, which is comparable to or better than most of the state of the art results~\Cite{chen2021lyapunov,li2020sample,qu2020finite} under Markovian observation models. Finite-time analysis of asynchronous Q-learning has been actively investigated recently in~\cite{chen2021lyapunov,li2020sample,qu2020finite} under Markovain observation models. Compared to the previous works, the main differences lie in conditions on the diminishing step-size rules. In particular, our conditions on the diminishing step-sizes are independent of the mixing time, where the mixing time is a quantity that characterizes the convergence speed of the current state-action distribution to the stationary distribution, and it plays an important role in the analysis of stochastic algorithms~\Cite{chen2021lyapunov,qu2020finite,li2020sample,doan2022finite,duchi2012ergodic} under Markovian noise scenarios. On the other hand, the previous works~\cite{chen2021lyapunov,li2020sample} employ conditions on step-sizes that depend on the mixing time or the covering time~\cite{beck2012error}, which is another quantity that characterizes the exploration performance. Therefore, the proposed new analysis covers different scenarios. Moreover, some of the previous approaches require deliberate processes to predict the mixing time of the underlying Markov chain~\Cite{wolfer2019estimating,hsu2015mixing}, which are not required in our approach.


 \item We revisit the control theoretic analysis in~\Cite{lee2021discrete}, and consider a diminishing step-size. With the diminishing step-size, it is hard to directly apply the proof techniques in~\Cite{lee2021discrete} due to time-varying nature of the diminishing step-size, which transfers the underlying linear switching system into a time-varying system, and it is in general much more challenging to analyze. To overcome this difficulty, we develop a new approach mainly based on a domain transformation~\Cite{lakshminarayanan2018linear} to a new coordinate, which facilitates analysis of convergence on the negative definite cone. Besides, we also take into account the last-time convergence and Markovian observation models in contrary to~\Cite{lee2021discrete}. Finally, we stress that our goal is to provide new insights and analysis framework to deepen our understanding on Q-learning via its unique connection to discrete-time switching systems, rather than improving existing convergence rates. In this respect, we view our analysis technique as a complement rather than a replacement for existing techniques for Q-learning.

 \end{enumerate}

\paragraph{Related Works}
Since Q-learning was developed by Watkins in~\Cite{watkins1992q}, several advances have been made in a series of works. Here, we summarize related results in the literature. 
\begin{enumerate} 
\item Asymptotic convergence: Convergence of synchronous Q-learning was analyzed for the first time in the original paper~\Cite{watkins1992q}. Subsequently, convergence of asynchronous Q-learning was studied in~\Cite{jaakkola1993convergence,tsitsiklis1994asynchronous} using results in stochastic approximation and contraction mapping property. Moreover, an ordinary differential equation (O.D.E) approach was developed in~\Cite{borkar2000ode} to prove general stochastic approximations, and applied it to prove convergence of synchronous Q-learning. More recently,~\Cite{lee2020unified} developed a switching system model of asynchronous Q-learning, and applied the O.D.E method for its convergence.

\item Non-asymptotic convergence with diminishing step-sizes: The early result in~\Cite{szepesvari1997asymptotic} considered state-dependent diminishing step-sizes to study non-asymptotic convergence under independent and identically distributed (i.i.d.) observation models. There are a few works focusing on the Markovian sampling setting together with diminishing step-sizes. In particular, \Cite{even2003learning} obtained sample complexity to achieve $\epsilon$-optimal solution with high probabilities by extending the convergence analysis of~\Cite{bertsekas1995neuro}. More recently, \Cite{qu2020finite} considered  a general asynchronous stochastic approximation scheme featuring a weighted infinity-norm contractive operator, and proved a bound on its finite-time convergence rate under the Markovian sampling setting. They used the notion of the so-called sufficient exploration and error decomposition with shifted martingale difference sequences. Another recent work in~\Cite{chen2021lyapunov} employed the so-called generalized Moreau envelope and the notion of Lyapunov function. The recent study in~\Cite{li2020sample} provided a probability tail bound through refined analysis based on an adaptive step-size, which depends on the number of visits to each state-action pair. Besides, \Cite{lee2020periodic} proposed an optimization-based Q-learning that mimics the recent deep Q-learning in~\Cite{mnih2015human}, and offered new finite-time convergence results.

\item Non-asymptotic convergence with constant step-sizes: The aforementioned works applied time-varying step-sizes such as diminishing or adaptive step-sizes. On the other hand, there are some works dealing with Q-learning under constant step-sizes. For instance, \Cite{beck2012error,beck2013improved} employed the concept of the so-called covering time, \(t_{\text{cover}}\), to derive mean squared error bounds. Using similar approaches as in the diminishing step-size analysis~\Cite{li2020sample}, \Cite{chen2021lyapunov} proposed new convergence analysis under Markovian observation models. More recently, \Cite{lee2021discrete} developed a discrete-time switching system model of asynchronous Q-learning to compute a convergence rate of averaged iterated under i.i.d. observation model.

\item Control system approach: Finally, it is worth summarizing related studies based on control system frameworks. Dynamical system perspectives of reinforcement learning and general stochastic iterative algorithms have a long tradition, and they date back to O.D.E analysis~\Cite{borkar2009stochastic,kushner2003stochastic,bhatnagar2012stochastic,borkar2000ode}. Recently,~\Cite{hu2019characterizing} studied asymptotic convergence of temporal difference learning (TD-learning)~\Cite{sutton1988learning} based on a Markovian jump linear system model. They tailored Markovian jump linear system theory developed in the control community to characterize exact behaviors of the first and second order moments of TD-learning algorithms. The paper~\cite{lee2020unified} studied asymptotic convergence of Q-learning~\Cite{watkins1992q} through a continuous-time switched linear system model~\Cite{liberzon2003switching}. A finite-time analysis of Q-learning was also investigated in~\Cite{lee2021discrete} using discrete-time switched linear system models.

\end{enumerate}

\par




The paper is organized as follows.
\begin{enumerate}
    \item \Cref{sec:prelim} provides preliminary backgrounds on Markov decision process, dynamic system, Q-learning, essential notations and definitions used throughout the paper.
    
    \item In~\Cref{sec:csa}, we introduce a dynamic system model of Q-learning. In particular, Q-learning is represented by a stochastic linear time-varying switching system with an affine input term. 
    \item \Cref{sec:ls} is devoted to analysis of the lower comparison system. We provide essential properties of the lower comparison system, and prove its convergence through a domain transformation under i.i.d. observation model. 
    
    \item In~\Cref{sec:us}, convergence of the upper comparison system is given under i.i.d. observation model.
    
    \item Based on the results in the previous two sections, the convergence of the lower and upper comparison systems,~\Cref{sec:os} provides a proof of the convergence of Q-learning under i.i.d. observation model.

    \item \Cref{sec:fta_markov} extends the analysis in the previous sections with i.i.d. scenario to Markovian observation case. 
    
    \item Finally, the overall results are concluded in~\Cref{sec:conclusion}.       
 
\end{enumerate}

%% file: preliminaries/notation.tex
The adopted notation throughout the paper is as follows: ${\mathbb R}$: set of real numbers; ${\mathbb R}^n $: $n$-dimensional Euclidean
space; ${\mathbb R}^{n \times m}$: set of all $n \times m$ real
matrices; $A^{\top}$: transpose of matrix $A$; $A \succ 0$ ($A \prec
0$, $A\succeq 0$, and $A\preceq 0$, respectively): symmetric
positive definite (negative definite, positive semi-definite, and negative semi-definite, respectively) matrix $A$; \( x \leq y \) and \( x \geq y \) for vectors $x,y$: element-wise inequalities; $I$: identity matrix with appropriate dimensions; $\rho(\cdot)$: spectral radius; for any matrix $A$, $[A]_{ij}$ is the element of $A$ in $i$-th row and $j$-th column; $\lambda_{\min}(A)$ and $\lambda_{\max}(A)$ for any symmetric matrix $A$: the minimum and maximum eigenvalues of $A$, respectively; \(\sigma_{\min}(A)\) and \(\sigma_{\max}(A)\): minimum and maximum singular value of matrix $A$, respectively; $|{\cal S}|$: cardinality of a finite set $\cal S$.

%% file: preliminaries/markov_decision_process.tex
In this paper, we consider an infinite horizon Markov decision process (MDP). It consists of a tuple \(\mathcal{M}= (\mathcal{S},\mathcal{A},P,r,\gamma)\), where \(\mathcal{S}\) and \(\mathcal{A}\) represent finite state and action spaces, respectively, \( P(s'|s,a) \) denotes the transition probability when action \(a\) is taken at state \(s\), and transition occurs to \(s'\), \( r: \mathcal{S} \times \mathcal{A} \times \mathcal{S} \rightarrow \mathbb{R}\) denotes the reward function, and \(\gamma \in (0,1) \) is the discount factor. 
With a stochastic policy \( \pi: \mathcal{S} \rightarrow \mathcal{P}(\mathcal{A})\), where $\mathcal{P}(\mathcal{A})$ is the set of probability distributions over $\mathcal A$, agent at the current state \(s\) selects an action \(a \sim \pi (\cdot|s) \), then the agent's state changes to the next state \( s' \sim P(\cdot | s,a) \), and receives reward \(r := r(s,a,s') \). For a deterministic policy \(\pi:\mathcal{S}\to\mathcal{A}\), \(\pi(s) \in \mathcal{A} \) is selected without randomness at state \(s\).

The objective of the Markov decision problem (MDP) is to find an optimal policy, denoted by $\pi^*$, such that the cumulative discounted rewards over infinite time horizons is maximized, i.e.,
\begin{align*}
\pi^*:= \argmax_{\pi} {\mathbb E}\left[ \left.\sum_{k=0}^\infty {\gamma^k r_k}\right|\pi\right],
\end{align*}
where $(s_0,a_0,r_0,s_1,a_1,r_1,\ldots)$ is a state-action trajectory generated by the Markov chain under policy $\pi$, $r_k$ denotes the reward received at time $k$ (will be detailed later), and ${\mathbb E}[\cdot|\pi]$ is an expectation conditioned on the policy $\pi$. 

The Q-function under policy $\pi$ is defined as
\begin{align*}
&Q^{\pi}(s,a)={\mathbb E}\left[ \left. \sum_{k=0}^\infty {\gamma^k r_k} \right|s_0=s,a_0=a,\pi \right],\quad  (s ,a)\in {\cal S} \times {\cal A},
\end{align*}
and the optimal Q-function is defined as $Q^*(s,a)=Q^{\pi^*}(s,a)$ for all $(s, a)\in {\cal S}\times {\cal A}$. Once $Q^*$ is obtained, then an optimal policy can be retrieved by the greedy action with respect to $Q^*$, i.e., $\pi^*(s)=\argmax_{a\in {\cal A}}Q^*(s,a)$. Throughout, we assume that the MDP is ergodic~\Cite{levin2017markov} so that the stationary state distribution exists, is unique, and the Markov decision problem is well posed.

It is known that the optimal Q-function satisfies Bellman equation expressed as follows:
\begin{align}
        &Q^*(s,a)\nonumber \\
        =& \mathbb{E}\left[\left.r_{k+1} + \max_{a_{k+1}\in \mathcal{A}} \gamma Q^{*}(s_{k+1},a_{k+1})\right|s_k = s,a_k = a\right]\label{eq:Bellman-equation}.
\end{align}
Finally, some notations related to MDP are summarized as follows: \(P_a \in \mathbb{R}^{|{\mathcal{S}}|\times|{\mathcal{S}}|}, a \in {\mathcal{A}}\) is the state transition matrix when action \(a\) is taken at state \(s\in \mathcal{S}\). Likewise, \(R_a\in \mathbb{R}^{|\mathcal{S}|},a\in\mathcal{A}\) is the reward vector when action \(a\) is taken at state \(s\in \mathcal{S}\), i.e., \(R_a(s):={\mathbb E}[r(s,a,s')|s,a], s\in {\mathcal{S}}\). Moreover, the notation \(Q_a:= Q(\cdot,a)\in {\mathbb R}^{|{\mathcal{S}}|},a\in {\mathcal{A}}\) will be adopted throughout the paper.

%% file: preliminaries/switching_system.tex
In this subsection, we briefly review the switching system, which will play an important role in this paper. The switching system is a special form of nonlinear systems, and hence, we first consider the nonlinear system
\begin{align}
x_{k+1}=f(x_k),\quad x_0=z \in {\mathbb R}^n,\quad k\in \{0,1,\ldots \},\label{eq:nonlinear-system}
\end{align}
where $x_k\in {\mathbb R}^n$ is the state and $f:{\mathbb R}^n \to {\mathbb R}^n$ is a nonlinear mapping. An important concept in dealing with the nonlinear system is the equilibrium point. A point $x=x^*$ in the state-space is said to be an equilibrium point of~\eqref{eq:nonlinear-system} if it has the property that whenever the state of the system starts at $x^*$, it will remain at $x^*$~\cite{khalil2002nonlinear}. For~\eqref{eq:nonlinear-system}, the equilibrium points are the real roots of the equation $f(x) = 0$. The equilibrium point $x^*$ is said to be globally asymptotically stable if for any initial state $x_0 \in {\mathbb R}^n$, $x_k \to x^*$ as $k \to \infty$.

Next, let us consider the particular nonlinear system, the \emph{linear switching system},
\begin{align}
&x_{k+1}=A_{\sigma_k} x_k,\quad x_0=z\in {\mathbb
R}^n,\quad k\in \{0,1,\ldots \},\label{eq:switched-system}
\end{align}
where $x_k \in {\mathbb R}^n$ is the state, $\sigma\in {\mathcal M}:=\{1,2,\ldots,M\}$ is called the mode, $\sigma_k \in
{\mathcal M}$ is called the switching signal, and $\{A_\sigma,\sigma\in {\mathcal M}\}$ are called the subsystem matrices. The switching signal can be either arbitrary or controlled by the user under a certain switching policy. Especially, a state-feedback switching policy is denoted by $\sigma_k = \sigma(x_k)$. A more general class of systems is the {\em affine switching system}
\begin{align*}
&x_{k+1}=A_{\sigma_k} x_k + b_{\sigma_k},\quad x_0=z\in {\mathbb
R}^n,\quad k\in \{0,1,\ldots \},
\end{align*}
where $b_{\sigma_k} \in {\mathbb R}^n$ is the additional input vector, which also switches according to $\sigma_k$. Due to the additional input $b_{\sigma_k}$, its stabilization becomes much more challenging. 

%% file: preliminaries/ltv.tex
In this section, we briefly review the linear time-varying (LTV) system~\Cite{zhou2017asymptotic}. 
Let us first consider the continuous-time linear time-invariant (LTI) system 
\begin{align}
    \dot{x}_t = A x_t, \label{eq:lti_system}
\end{align}
where \( x_t \in \mathbb{R}^n\) is the state, \( A \in \mathbb{R}^{n\times n}\) is the system matrix. The asymptotic stability of~\eqref{eq:lti_system} is completely characterized by the notion called Hurwitz matrix~\Cite{khalil2002nonlinear,chen2004linear}, which is defined as follows.  
\begin{definition}[Hurwitz matrix~\Cite{khalil2002nonlinear,chen2004linear}] A square matrix
\( A \in \mathbb{R}^{n\times n}\) is called Hurwitz if its real part of eigenvalues are negative. 
\end{definition}
It is well-known that~\eqref{eq:lti_system} is asymptotically stable if and only if $A$ is Hurwitz. In general, it is hard to check the Hurwitzness without computing the eigenvalues. Nevertheless, a sufficient condition can be obtained in terms of algebraic relations among entries of the corresponding matrix, which is called the strict diagonal dominance.
\begin{definition}[Strict diagonal dominance~\Cite{horn2013matrix}]\label{def:sdd}
A square matrix \(A  \in \mathbb{R}^{n\times n}\) is said to be strictly diagonally dominant if 
\begin{align*}
 |[A]_{ii}|>  \sum\limits_{j \ne i,j = 1}^n |[A]_{ij}|, \quad \forall{i}\in\{1,2,\dots,n\}.
\end{align*} 
\end{definition}

\begin{lemma}\label{lem:diag_dom_hurwtiz}
A square matrix $A$ is Hurwitz if it is a strictly row diagonally dominant matrix and the diagonal elements are negative. 
\end{lemma}
The proof of~\Cref{lem:diag_dom_hurwtiz} is in Appendix~\Cref{app:lem:diag_dom_hurwtiz}.

A linear time-varying (LTV) system is a class of linear systems defined as
\begin{align}
    x_{k+1} = A_k x_k, \label{eq:ltv_system}
\end{align}
where \( x_k \in \mathbb{R}^n\) is the state, and \( A_k \in \mathbb{R}^{n\times n}\) is the time-varying system matrix. Together with switching systems, LTV system models will be used in the analysis of this paper. Stability analysis of LTV systems is much more challenging than a simple linear time-invariant system~\Cite{chen2004linear}. For their asymptotic stability, the following sufficient condition will be used in this paper.
\begin{lemma}[\Cite{zhou2017asymptotic}, Theorem 1]\label{lem:ltv_stability}
The discrete-time LTV system~\eqref{eq:ltv_system} is asymptotically stable if there exists a sequence of positive definite matrices \( \{P_k\}_{k=0}^{n} \), and positive scalars \(\{p_k\}_{k=0}^{n}\) such that the following conditions hold for all \(k \geq 0\): 
\begin{enumerate}[1)]
 \item  \( A^{\top}_k P_{k+1}A_k \preceq p_k P_k \); 
  \item \(       \Pi_{k=1}^{n} p_k \to 0 \) as $n \to \infty$;
  \item \( c \leq \lambda_{\min}(P_k)\) for some positive constant \(c>0\).
\end{enumerate}
\end{lemma}
\begin{proof}
Let \( V_k (x_k) = x^{\top}_k P_k x_k \) be a time-varying Lyapunov function that satisfies the condition in item~1). Recursively applying the inequality in item~1) leads to
\begin{align*}
V_{k}(x_{k}) \leq p_{k-1} V_{k-1} (x_{k-1})\leq \left( \prod\nolimits_{i = 0}^{k - 1} {p_i }  \right) V_0(x_0).
\end{align*}
Using item~3), it follows from the last inequality that
\begin{align*}
c\left\| {x_k } \right\|_2^2  \le V_k (x_k ) \le \left( \prod\nolimits_{i = 0}^{k - 1} {p_i }  \right)V_0 (x_0 ) \le \left( {\prod\nolimits_{i = 0}^{k - 1} {p_i } } \right)\lambda _{\max } (P_0 ),
\end{align*}
which leads to 
\begin{align*}
\|x_k\|_2^2  \le \left( {\prod\nolimits_{i = 0}^{k - 1} {p_i } } \right)\frac{{\lambda _{\max } (P_0 )}}{c}.
\end{align*}
By item~2), we have \( \|x_k\|_2 \to 0 \) as $k \to \infty$. This completes the proof.
\end{proof}


%% file: preliminaries/qlearning_notations.tex
In this paper, a linear time-varying switching system model of Q-learning will be used for the analysis. To this end, the following matrix notations will be useful to compactly represent Q-learning into a vector and matrix forms:
\begin{align*}
P:= \begin{bmatrix}
   P_1\\
   \vdots\\
   P_{|{\mathcal{A}}|}
\end{bmatrix} \in {\mathbb R}^{ |{\mathcal{S}}||{\mathcal{A}}|\times |{\mathcal{S}}| } ,\quad R:= \begin{bmatrix}
   R_1 \\
   \vdots \\
   R_{|{\mathcal{A}}|} 
\end{bmatrix} \in {\mathbb R}^{|{\mathcal{S}}||{\mathcal{A}}|},\quad Q:= \begin{bmatrix}
   Q_1\\
  \vdots\\
   Q_{|{\mathcal{A}}|}\\
\end{bmatrix}\in {\mathbb R}^{|{\mathcal{S}}||{\mathcal{A}}|}.
\end{align*}
Under policy \(\pi\), we have \(P^{\pi}\in \mathbb{R}^{|{\mathcal{S}}||{\mathcal{A}}|\times|{\mathcal{S}}||{\mathcal{A}}|}\) as the state-action transition matrix, i.e.,  
\begin{align*}
&(e_s  \otimes e_a )^{\top} P^\pi  (e_{s'}  \otimes e_{a'} )= {\mathbb P}[s_{t + 1}  = s',a_{t + 1}  = a'|s_t  = s,a_t  = a,\pi],
\end{align*}
where $\otimes$ stands for the Kronecker product, and \( e_a \) and \( e_s \) to represent \(a\)-th and \(s\)-th canonical basis vectors in \(\mathbb{R}^{|\mathcal{A}|}\) and  \(\mathbb{R}^{|\mathcal{S}|}\), respectively.

In this paper, we represent a policy in a matrix form. In particular, for a given policy $\pi$, define the matrix
\begin{align}
    \Pi^{\pi} := 
    \begin{bmatrix}
    (e_{\pi(1)} \otimes e_1)^{\top}\\
    \vdots\\
    (e_{\pi(|\mathcal{S}|)} \otimes e_{|\mathcal{S}|})^{\top}\\
    \end{bmatrix}
    \in \mathbb{R}^{|\mathcal{S}|\times |\mathcal{S}||\mathcal{A}|} . 
\end{align}
Then, we can prove that for any deterministic policy, $\pi$, we have 
\begin{align*}
\Pi^\pi  Q = \begin{bmatrix}
   {Q(1,\pi (1))}  \\
   {Q(2,\pi (2))}  \\
    \vdots   \\
   {Q(|{\mathcal{S}}|,\pi (|{\mathcal{S}}|))}  \\
\end{bmatrix}.
\end{align*}

With this \(\Pi^{\pi}\), we can prove that for any deterministic policy $\pi$, \( P^{\pi} = P\Pi^{\pi} \in \mathbb{R}^{ |\mathcal{S}||\mathcal{A}|\times  |\mathcal{S}||\mathcal{A}|}\), where $P^{\pi}$ stands for the state-action transition matrix. Using the notations introduced, the Bellman equation in~\eqref{eq:bellman_optimal_q} can be compactly written as 
\begin{equation}\label{eq:bellman_optimal_q}
Q^*  = \gamma P\Pi _{\pi _{Q^* } } Q^*  + R ,
\end{equation}
where $\pi_{Q^*}$ is the greedy policy defined as $\pi_{Q^*} (s) = \argmax_{a\in \mathcal{A}} Q^*(s,a)$. Moreover, for simplicity of notations, we introduce the following shorthand: 
\begin{align*}
\Pi_{Q} := \Pi^{\pi_Q},
\end{align*}
when the policy is the greedy policy $\pi_{Q} (s) = \argmax_{a\in \mathcal{A}} Q(s,a)$ with respect to $Q$.

%% file: preliminaries/revisit_qlearning.tex
This section briefly reviews standard Q-learning and its important properties. Two versions of Q-learning will be investigated in this paper. 
\begin{algorithm}[h!]
\caption{Q-learning with i.i.d. observation model}
  \begin{algorithmic}[1]
    \State Initialize $Q_0 \in {\mathbb R}^{|{\cal S}||{\cal A}|}$.
    \State Set the step-size $(\alpha _k )_{k = 0}^\infty$, and the behavior policy $\beta$.
    \For{iteration $k=0,1,\ldots$}
        \State Sample $s_k\sim p$ and $a_k \sim \beta$
        \State Sample $s_k'\sim P(s_k,a_k,\cdot)$ and $r_k= r(s_k,a_k,s_k')$
        \State Update $Q_{k+1}(s_k,a_k)=Q_k(s_k,a_k)+\alpha_k \{r_k+\gamma\max_{u \in {\cal A}} Q_k(s_k',u)-Q_k (s_k,a_k)\}$
    \EndFor
  \end{algorithmic}\label{algo:standard-Q-learning1}
\end{algorithm}
\begin{algorithm}[h!]
\caption{Q-learning with Markovian observation model}
  \begin{algorithmic}[1]
    \State Initialize $Q_0 \in {\mathbb R}^{|{\cal S}||{\cal A}|}$.\\
    \State Set the step-size $(\alpha _k )_{k = 0}^\infty$, and the behavior policy $\beta$.
    \State Sample $s_0\sim p_0$
    \For{iteration $k=0,1,\ldots$}
        \State Sample $a_k \sim \beta$
        \State Sample $s_{k+1}\sim P(s_k,a_k,\cdot)$ and $r_k= r(s_k,a_k,s_{k+1})$
        \State Update $Q_{k+1}(s_k,a_k)=Q_k(s_k,a_k)+\alpha_k \{r_k+\gamma\max_{u \in {\cal A}} Q_k(s_{k+1},u)-Q_k (s_k,a_k)\}$
    \EndFor
  \end{algorithmic}\label{algo:standard-Q-learning2}
\end{algorithm}
\cref{algo:standard-Q-learning1} is Q-learning with i.i.d. observation model, while \cref{algo:standard-Q-learning2} is Q-learning with Markovian observation model. In~\cref{algo:standard-Q-learning1}, $p$ denotes the current state distribution, and $\beta$ is the so-called behavior policy, which is the policy to collect sample trajectories. On the other hand, in~\cref{algo:standard-Q-learning2}, $p_0$ stands for the initial state distribution. Under the i.i.d. sampling setting in~\cref{algo:standard-Q-learning1}, at each iteration $k$, single state-action tuple \( (s_k,a_k) \) is sampled from the distribution \(d\) defined by
\[
d(s,a) = {\mathbb P}[s_k = s,a_k = a] = p(s)\beta (a|s),\quad \forall (s,a) \in {\cal S} \times {\cal A}.
\]
The next state \( s_k'\) is sampled independently at every iteration following the state transition probability $P(s_k,a_k,\cdot)$. Note that this is similar to the generative setting~\Cite{li2021tightening}, which assumes that every state-action tuple is generated at once from an oracle. With a slight abuse of notation, $d$ will be also used to denote the vector $d \in {\mathbb R}^{|{\mathcal{S}}||{\mathcal{A}}|}$ such that
\[
d^{\top} (e_s  \otimes e_a ) = d(s,a),\quad \forall (s,a) \in \mathcal{S} \times \mathcal{A}.
\]
Moreover, we further adopt the following notations:
\begin{align*}
    d_{\max} := \max_{(s,a)\in\mathcal{S}\times\mathcal{A}}d(s,a),\quad d_{\min} := \min_{(s,a)\in\mathcal{S}\times\mathcal{A}} d(s,a). 
\end{align*}
and
\begin{align*}
r_k : = r(s_k ,a_k ,s_k ').
\end{align*}

On the other hand, Q-learning with Markovian observation model in~\cref{algo:standard-Q-learning2} uses samples from a single state-action trajectory which follows the underlying Markov chain with the state-action transition matrix $P^\beta$, where $\beta$ is the so-called behavior policy. To be more specific, we suppose that starting from an initial state-action tuple $(s_0,a_0)\in {\cal S} \times {\cal A}$ with the initial distribution $\mu_0$, the underlying MDP generates $(s_0,a_0,r_0,s_1,a_1,r_1,\ldots)$ under the behavior policy $\beta$ so that the current state-action distribution $\mu_k$ is defined as
\begin{equation}\label{eq:mu_k}
\mu _k(s,a) : = {\mathbb P}[s_k=s ,a_k = a|\beta ],\quad \forall (s,a) \in {\cal S} \times {\cal A}.
\end{equation}
with the initial state-action distribution $\mu_0$ determined by
\[
\mu _0 (s,a) = {\mathbb P}[s_0  = s,a_0  = a] = p_0 (s)\beta (a|s),\quad \forall (s,a) \in {\cal S} \times {\cal A}.
\]

With the fixed behavior policy $\beta$, the MDP becomes a Markov chain with the state-action transition probability matrix $P^\beta:=P\Pi^\beta$, i.e., 
\[
\mu _0^{\top} (P^\beta)^k   = \mu _k^{\top}.
\]
Throughout the paper, we adopt the following standard assumption on this Markov chain.
\begin{assumption}\label{assmp:irreducible_aperiodic} Consider the Markov chain of the state-action pair $(s,a) \in {\cal S} \times {\cal A}$ under $\beta$ and initial state distribution $p_0$. We assume that the Markov chain is aperiodic and irreducible~\Cite{levin2017markov}. 
\end{assumption}
This assumption implies that under $\beta$, the corresponding Markov chain admits the unique stationary distribution defined as
\begin{equation}\label{eq:mu_k_infty}
\mu _\infty ^{\top} : = \mathop {\lim }\limits_{k \to \infty } \mu _k^{\top}  = \mathop {\lim }\limits_{k \to \infty } \mu _0^{\top} (P^\beta  )^k .
\end{equation}

In the Markovian observation case in~\cref{algo:standard-Q-learning2}, \(d_{\max}\) and \(d_{\min}\) are defined as 
\begin{align*}
d_{\max} := \max_{(s,a)\in\mathcal{S}\times\mathcal{A}}\mu_\infty(s,a),\quad d_{\min} := \min_{(s,a)\in\mathcal{S}\times\mathcal{A}} \mu_\infty(s,a).
\end{align*}

Moreover, the current reward $r_k$ is defined as
\begin{align*}
r_k : = r(s_k ,a_k ,s_{k+1}).
\end{align*}

To analyze the stochastic process generated by Q-learning, we define the notation \( \mathcal{F}_k\) as follows:
\begin{enumerate}
    \item \( \mathcal{F}_k\) for i.i.d. observation: \(\sigma\)-field induced by \(\{ (s_i,a_i,s_{i}',r_i, Q_i)\}_{i=0}^k\)

    \item \( \mathcal{F}_k\) for Markovian observation: \(\sigma\)-field induced by \(\{ (s_i,a_i,s_{i+1},r_i, Q_i)\}_{i=0}^k\)
\end{enumerate}

In this paper, we will consider i.i.d. observation model first, and then the corresponding analysis will be extended to the Markovian observation model later. 
Next, for simplicity of the proof, we assume the following bounds on the initial value, \(Q_0\) and the reward matrix \(R\).
\begin{assumption}\label{assm:Q_0_reward_bound}
Throughout the paper, we assume that the initial value, \( Q_0\), and reward, \(R\), is bounded as follows:
\begin{align*}
||Q_0||_{\infty} \leq 1 ,\quad ||R||_{\infty} \leq 1.
\end{align*}
\end{assumption}

With~\Cref{assm:Q_0_reward_bound}, we can derive bounds on \(Q_k\) and \(w_k\), which will be useful throughout the paper. First, we establish a bound on \(Q_k\).
\begin{lemma}\label{lem:q_k_bound}
For all \(k \geq 0\), we have 
\begin{align}
 ||Q_k||_{\infty} \leq \frac{1}{1-\gamma},\quad \forall k \geq 0. \nonumber
\end{align}
\end{lemma}
The proof is in Appendix~\Cref{app:proof:q_k_bound}.

%% file: preliminaries/linear_sa.tex
Linear stochastic approximation~\Cite{robbins1951stochastic} is widely used to find a fixed point \(x^* \) of linear system, \( Ax = b \), where \( A \in \mathbb{R}^{m \times m}\), and $b\in \mathbb{R}^{m\times 1}$ are constants, $x\in \mathbb{R}^{m\times 1}$ is an unknown vector. Suppose that a direct access to \(A\) and \(b\) is not available, while their stochastic estimations are accessible. 
The linear stochastic approximation is the following update: 
\begin{align}
    x_{k+1} = x_k + \alpha_k ( A x_k -b + w_k ) \label{eq:general_sa},
\end{align}
where \(\alpha_k \) is the step-size, and $w_k \in \mathbb{R}^{m\times 1}$ is an i.i.d. noise with zero mean, which is made due to the incomplete information on $(A,b)$. Under some mild assumptions, the stochastic recursion is known to converge to a fixed point, $x^*$, with probability one. When the noises \( (w_k)_{k=1} \) are Markovian, the convergence analysis of~\cref{eq:general_sa} is more challenging. To avoid this issue, the notion of uniform ergodicity of Markov chain has been introduced in~\Cite{levin2017markov}. Further details on the Markovian sampling will be discussed later in~\Cref{sec:fta_markov}.

%% file: preliminaries/control_system_approach.tex
In this section, we reformulate Q-learning update in~\cref{algo:standard-Q-learning1} into a state-space representation of a control system~\Cite{khalil2002nonlinear}, which is commonly used in control literature. Using matrix notations, the Q-learning algorithm with i.i.d. observation model in~\cref{algo:standard-Q-learning1} can be written as follows:
\begin{equation}\label{eq:q_learning_sa}
    Q_{k+1} = Q_k +\alpha_k (DR + \gamma DP\Pi_{Q_k}-DQ_k + w_k) ,
\end{equation}
where $D \in {\mathbb R}^{|{\cal S}||{\cal A}| \times |{\cal S}||{\cal A}|}$ is a diagonal matrix with diagonal elements being the current state-action distributions, i.e., 
\[
(e_s  \times e_a )^{\top} D (e_s  \times e_a ) = d(s,a) = {\mathbb P}[s_k  = s,a_k  = a],
\]
and
\begin{align}
    w_k &:= (e_{a_k} \otimes e_{s_k} ) (e_{a_k} \otimes e_{s_k})^{\top} R + \gamma (e_{a_k} \otimes e_{s_k})(e_{s_k'})^{\top}\Pi_{Q_k}Q_k \label{eq:noise_term}\\
    &- (e_{a_k} \otimes e_{s_k})(e_{a_k}\otimes e_{s_k})^{\top} Q_k - (DR+ \gamma DP\Pi_{Q_k}Q_k - DQ_k) \nonumber
\end{align}
is an i.i.d. noise term. Note that the noise term is unbiased conditioned on $Q_k$, i.e., ${\mathbb E}[w_k|Q_k] = 0$. 
Moreover, using the above bound on \(||Q_k||_{\infty}\), a bound on \(w_k\) can be established as follows.
\begin{lemma}\label{lem:w_k_bound}
Under i.i.d. observation model, the stochastic noise, \(w_k \), defined in~\eqref{eq:noise_term}, is bounded as follows:
\begin{align}
    ||w_k||_{\infty} &\leq \frac{4}{1-\gamma} , \label{ineq:bound on w_k} \\
     \mathbb{E}[||w_k||^2_2 | \mathcal{F}_k] &\leq \frac{9}{(1-\gamma)^2} \label{ienq:bound_w_k_2} 
\end{align}
for all \(k \geq 0 \), where \( \mathcal{F}_k\) is a \(\sigma\)-field induced by \(\{ (s_i,a_i,s_{i}',r_i, Q_i)\}_{i=0}^k\). 
\end{lemma}

The proof is given in Appendix~\Cref{app:proof:w_k_bound}. Furthermore, using Bellman equation in~\eqref{eq:bellman_optimal_q}, one can readily prove that the error term \(Q_k - Q^* \) evolves according to the following dynamic system:
\begin{align}
 Q_{k+1}-Q^* = A_{Q_k,k}(Q_k - Q^*)+b_{Q_k,k}+\alpha_k w_k, \label{eq:simplified_qlearning_sa}
\end{align}
where
\begin{align}\label{def:A,b}
    A_{Q,k} := I+\alpha_k T_Q ,\quad b_{Q,k}:= \alpha_k  \gamma DP(\Pi_Q - \Pi_{Q^*})Q^*,\quad T_{Q} := \gamma DP\Pi_{Q}-D .
\end{align}
In particular, this representation follows from the relations 
\begin{align*}
        Q_{k+1}-Q^* &= Q_k -Q^* +\alpha_k (DQ^* - \gamma DP\Pi_{Q^*} Q^* + \gamma DP\Pi_{Q_k}Q_k -DQ_k + w_k) \nonumber \\
                    &= Q_k -Q^*+\alpha_k\{ \gamma DP\Pi_{Q_k}(Q_k - Q^*) +\gamma DP\Pi_{Q_k} Q^*\nonumber \\
                    &-\gamma DP\Pi_{Q^*}Q^* - D(Q_k - Q^*)+w_k\} \nonumber \\
                    &= Q_k -Q^* +\alpha_k\{(\gamma DP\Pi_{Q_k}-D)(Q_k - Q^*) \\
                    &+ (\gamma DP\Pi_{Q_k}-\gamma DP\Pi_{Q^*})Q^* + w_k \} \nonumber \\
                    &= A_{Q_k,k}(Q_k - Q^*)+b_{Q_k,k}+\alpha_k w_k, \nonumber
\end{align*}
where the first equality comes from the Bellman equation~\eqref{eq:bellman_optimal_q}, the second and third equalities are obtained by simply rearranging terms, and the last equality comes from~\eqref{def:A,b} just by simplifying the  related terms. Note that compared to the linear stochastic approximation, the Q-learning update, has affine term \( b_{Q,k} \), which imposes difficulty in the analysis of the algorithm.

Now, we can view \(A_{Q,k}\) as a system matrix of a stochastic linear time-varying switching system in the following senses:
\begin{enumerate} 
    \item The system matrix is switched due to the max operator;
    
    \item The system matrix is time-varying due to the diminising step-size. 

\end{enumerate}
In this paper, we will study convergence of Q-learning based on control system perspectives with the above interpretations.  In particular, to analyze the convergence, we will use notions in control community such as Lyapunov theory for stability analysis of dynamical systems. First of all, we introduce the following result.
\begin{lemma}[\Cite{lee2021discrete}]\label{lem:spectral_radius_of_A}
For any \( Q\in \mathbb{R}^{|\mathcal{S}||\mathcal{A}|} \), and \(k \geq 0\),
 \begin{equation}
     ||A_{Q,k}||_{\infty} \leq 1-\alpha_k d_{\min}(1-\gamma). \nonumber
 \end{equation}
\end{lemma}

\begin{proof}
The column sum for each row becomes
    \begin{align}
        \sum\limits_{j=1}^{|\mathcal{S}||\mathcal{A}|} |  [A_{Q,k}]_{ij} | &= \sum\limits^{|\mathcal{S}||\mathcal{A}|}_{j=1} | [I-\alpha_k D + \alpha_k \gamma DP\Pi_Q ]_{ij} |  \nonumber \\
                                          &= [I-\alpha_k D]_{ii} + \sum\limits_{j=1}^{|\mathcal{S}||\mathcal{A}|} [\alpha_k \gamma DP\Pi_Q]_{ij} \nonumber \\
                                          &= 1- \alpha_k [D]_{ii} + \alpha_k \gamma [D]_{ii} \sum\limits_{j=1}^{|\mathcal{S}||\mathcal{A}|} [P\Pi_Q]_{ij} \nonumber \\
                                          &= 1 -\alpha_k[D]_{ii} + \alpha_k \gamma [D]_{ii}, \nonumber
    \end{align}
where the second equality is due to the fact that each element of \(A_{Q,k}\) is positive. Taking maximum over the row index, we get
    \begin{align*}
        ||A_{Q,k}||_{\infty} &= \max_{i\in\{1,2,\dots,|\mathcal{S}||\mathcal{A}|\}} \left(1 + \alpha_k [D]_{ii} (\gamma -1)\right) \\
                             &= 1 -\alpha_k \min_{(s,a)\in \mathcal{S}\times\mathcal{A}} d(s,a) (1-\gamma ) .
    \end{align*}
\end{proof}
Using similar arguments, one can also prove that \( T_{Q}\), which can be thought of as system matrix for continuous dynamics, is Hurwitz matrix, which can be proved using the strict diagonal dominance in~\Cref{def:sdd}. 
\begin{lemma}\label{lem:T_q_hurwitz}
\(T_{Q}\) is Hurwitz for all \(Q\in \mathbb{R}^{|{\cal S}||{\cal A}|}\). In addition, we have \begin{align*}
    ||T_Q||_{\infty} \leq 2d_{\max} .
\end{align*}
\end{lemma}

\begin{proof}
From~\Cref{lem:diag_dom_hurwtiz}, it suffices to prove that \(T_{Q}\) is strictly row diagonally dominant which is defined in~\Cref{def:sdd}. Each component can be calculated as follows:
\begin{align}
    [T_{Q}]_{ii} &= [D]_{ii}(-1+\gamma e_i^T P\Pi_Q e_i)  \nonumber, \\
   \sum\limits_{j \ne i,j = 1}^{|\mathcal{S}||\mathcal{A}|}[T_{Q}]_{ij} &=\sum\limits_{j \ne i,j = 1}^{|\mathcal{S}||\mathcal{A}|} \gamma [D]_{ii}e^{\top}_iP\Pi_Q e_j = \gamma [D]_{ii}(1-e^{\top}_iP\Pi_{Q}e_i)  \nonumber.
\end{align}
Now, we can prove that the diagonal part dominates the off-diagonal part, i.e.,
\begin{align*}
|[T_{Q}]_{ii}| >\sum\limits_{j \ne i,j = 1}^{|\mathcal{S}||\mathcal{A}|}|[T_{Q}]_{ij}|,
\end{align*}
from the following relation:
\begin{align*}
   |[T_{Q}]_{ii}| - \sum\limits_{j \ne i,j = 1}^{|\mathcal{S}||\mathcal{A}|} | [T_{Q}]_{ij} |&=  [D]_{ii}(1-\gamma e^T_i P\Pi_Qe_i) - \gamma [D]_{ii}(1-e^{\top}_iP\Pi_{Q}e_i)\\
   &= (1-\gamma ) [D]_{ii} > 0  .
\end{align*}
Since the diagonal term is negative, \(T_Q\) is Hurwitz by~\Cref{lem:diag_dom_hurwtiz}. Moreover,
\begin{align*}
    ||T_{Q}||_{\infty} &= \max_{i\in\{1,2,\dots,|\mathcal{S}||\mathcal{A}|\}} \sum\limits^{|\mathcal{S}||\mathcal{A}|}_{j=1} |[T_Q]_{ij} | \\ &=\max_{i\in\{1,2,\dots,|\mathcal{S}||\mathcal{A}|\}} \{ (1+\gamma - 2\gamma e_i^{\top}P\Pi_{Q}e_i)[D]_{ii}\} \nonumber \\
    &\leq 2d_{\max} \nonumber.
\end{align*}
\end{proof}

\begin{figure*}[h!]
\centering
\includegraphics[width=10cm]{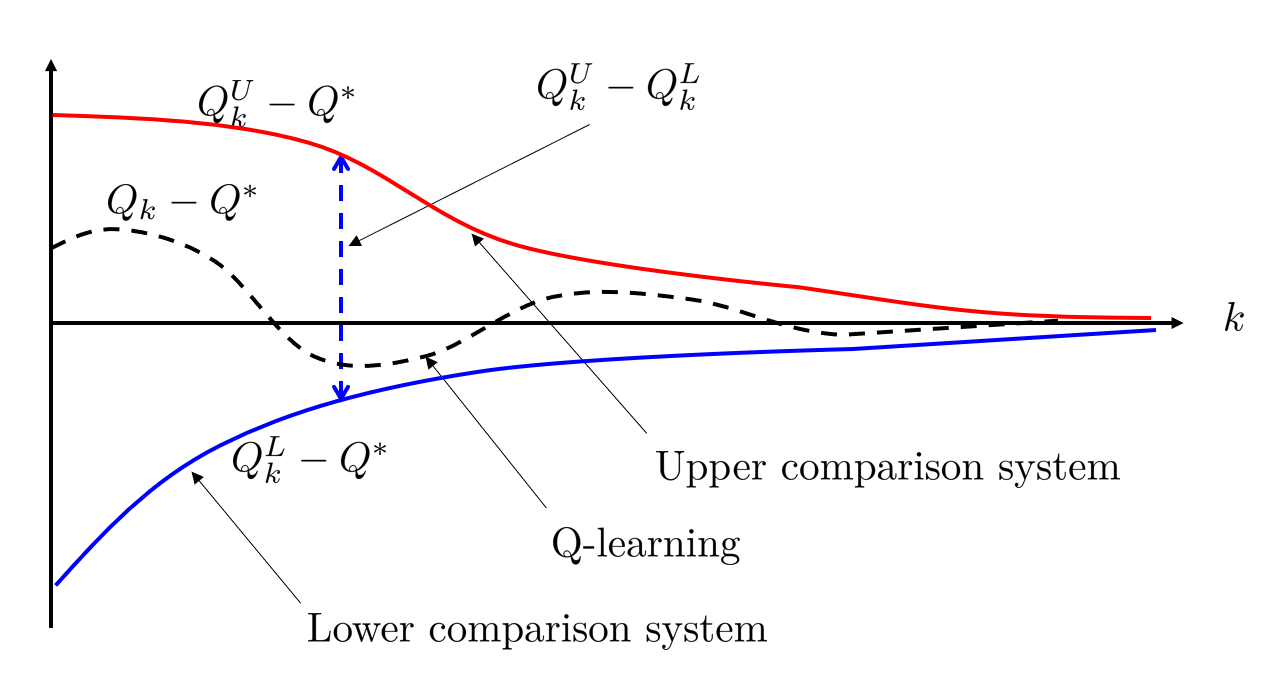}
\caption{Upper and lower comparison systems}\label{fig:1}
\end{figure*}

Proceeding on, consider the stochastic update~\eqref{eq:simplified_qlearning_sa}. Ignoring the stochastic term  \(\alpha_k w_k \), it can be viewed as a discrete-time switched affine system. Unfortunately, analysis of a switched affine system is more involved than that of a switched linear system. The recent result in~\Cite{lee2020unified} overcame this difficulty by introducing the notion of the upper and lower comparison systems corresponding to the original system. In particular,~\Cite{lee2020unified} introduced a switching system representation of the continuous-time O.D.E. model corresponding to Q-learning. Then, the stability has been proved based on comparison of the upper and lower comparison systems of the original system in~\eqref{eq:simplified_qlearning_sa}. Moreover, \Cite{lee2021discrete} extended this framework to a discrete-time setting, and derived finite-time convergence analysis under constant step-sizes. The overall idea is illustrated in~\Cref{fig:1}. However, when a time-varying step-size is used, it imposes significant difficulties in the analysis, which will be illustrated in the following sections. 

%% file: lower_system/header.tex
This section provides construction of a lower comparison system, and its finite time analysis. The construction of the lower comparison system follows that of~\Cite{lee2021discrete}. With the help of a similarity transform, the convergence analysis follows the spirit of~\Cite{nemirovski2009robust}, which is a well-known approach to prove convergence rate in optimization literature.

%% file: lower_system/construction.tex
By the lower comparison system, we mean a dynamic system whose state at time $k$, denoted by \(Q^L_k\), bounds the original system's state $Q_k$ from below on the orthant, i.e., \(Q^L_k \leq Q_k\). In particular, the lower comparison system can be expressed as follows: 
\begin{align}
    Q^L_{k+1} - Q^* = A_{Q^*,k}(Q^L_k - Q^*) + \alpha_k w_k. \label{eq:lower_comparison_update}
\end{align} 
Compared to the original system in~\eqref{eq:simplified_qlearning_sa}, the main difference is the affine term $b_{Q_k,k}$, which does not appear in the lower comparison system. For completeness, we provide the proof of the essential property, \(Q^L_k \leq Q_k\), in the following lemma.
\begin{lemma}[\Cite{lee2021discrete}]
 When \( Q^L_0 - Q^* \leq Q_0 - Q^* \), 
 \begin{align}
     Q^L_k - Q^* \leq Q_k - Q^* .\label{ineq:lower_comparison_iterate}
 \end{align}
\end{lemma}

\begin{proof}
The proof is done by induction. Assume for some \(k \geq 0\),~\eqref{ineq:lower_comparison_iterate} satisfies. Now, we prove that the inequality is also satisfied for \((k+1)\)-th iterate.
 \begin{align}
     Q^L_{k+1} - Q^* &= A_{Q^*,k} (Q^L_k - Q^*) + \alpha_k w_k \nonumber \\
                     &\leq A_{Q^*,k} (Q_k - Q^*) + \alpha_k w_k \nonumber \\
                     &= A_{Q_k,k} (Q_k - Q^*) -(  A_{Q_k,k} - A_{Q^*,k})(Q_k - Q^*) + \alpha_k w_k  \nonumber\\
                     &= A_{Q_k,k} (Q_k - Q^*) + \alpha_k ( \gamma DP\Pi_{Q^*}-\gamma DP\Pi_{Q_k})(Q_k-Q^*) + \alpha_k w_k \nonumber \\
                     &\leq A_{Q_k,k} (Q_k - Q^*) + \alpha_k ( \gamma DP\Pi_{Q_k}-\gamma DP\Pi_{Q^*})Q^* +\alpha_k w_k \nonumber\\
                     &= Q_{k+1} - Q^*, \nonumber
 \end{align}
 where the first inequality is due to the hypothesis and the fact that \( A_{Q,k}\) is positive matrix for any \(Q\). Moreover, the second inequality can be driven using the fact that \(  \Pi_{Q^*} Q_k \leq \Pi_{Q_k}Q_k\) and \(\Pi_{Q_k} Q^* \leq \Pi_{Q^*}Q^*\).
\end{proof}

To analyze the stochastic process generated by the lower comparison system, we define the notation \( \mathcal{F}_k^L\) as follows:
\begin{enumerate}
    \item \( \mathcal{F}_k^L\) for i.i.d. observation: \(\sigma\)-field induced by \(\{ (s_i,a_i,s_{i}',r_i, Q_i,Q^L_i)\}_{i=0}^k\)

    \item \( \mathcal{F}_k^L\) for Markovian observation: \(\sigma\)-field induced by \(\{ (s_i,a_i,s_{i+1},r_i, Q_i,Q^L_i)\}_{i=0}^k\)
\end{enumerate}

%% file: lower_system/lower_system.tex

This section provides a convergence rate of the lower comparison system. Let us consider the deterministic version of~\eqref{eq:lower_comparison_update} as follows:
\begin{align}
    x_{k+1} = A_{Q^*,k} x_k.\label{eq:deterministic_version}
\end{align}
 The above system can be seen as a linear time-varying system. From~\Cref{lem:ltv_stability}, the existence of a positive definite matrix \(P\) and a sequence \(\{p_k\} \) such that \( \Pi_{k=0}^{\infty}p_k \to 0 \) ensures asymptotic stability  of~\eqref{eq:deterministic_version}, when \(P\) and \( \{p_k\} \) satisfies below condition:
 \begin{align}\label{ineq:dissipativity}
      A^{\top}_{Q^*,k}PA_{Q^*,k} \preceq p_k P  \quad \forall{k} \geq 0,
 \end{align}
Here, \(p_k\) can be controlled by an appropriate step-size. To proceed, let us choose \(P = I \) to have 
\begin{align}
   A^{\top}_{Q^*,k}PA_{Q^*,k} - p_k P = \left((1-p_k)I+\alpha_k(T_{Q^*}+T_{Q^*}^{\top})+\alpha_k^2 T^{\top}_{Q^*}T_{Q^*} \right) \label{eq:1}.
\end{align}
We want to find a step-size $\alpha_k$ such that the above equation is negative semidefinite. To this end, it is necessary that the second term $T_{Q^*}+T_{Q^*}^{\top}$ is negative definite. However, since \(T_{Q^*}+T_{Q^*}^{\top}\) is not necessarily negative definite in general, it is hard to specify general conditions for a diminishing step-size to satisfy~\eqref{ineq:dissipativity}. To overcome this difficulty, we will consider the recent result in~\cite{lakshminarayanan2018linear}. In particular, \cite{lakshminarayanan2018linear} studied convergence of linear stochastic recursions in~\Cref{sec:lsa}, and a similarity transform was used in order to transform the matrix $A$ in~\eqref{eq:general_sa} into a negative definite matrix. In this paper, we will apply a similar idea to derive the convergence rate of the lower comparison system. First of all, we will find a similarity transform of $T_{Q^*}$ to a negative definite matrix using the fact that \(T_{Q^*}\) is Hurwitz.
Then, the Lyapunov theory can be applied to derive the convergence results. 


\subsubsection{Similarity transformation to a negative definite matrix}

In this subsection, we find a similarity transform of \(T_{Q^*}\) to a negative definite matrix. Such a similarity transform has been used in estimation of an upper bound of the solution of Lyapunov equation in~\Cite{fang1997new}. Moreover, in control community, a common similarity transform to an upper triangular matrix has been used to study a convergence rate of switched linear systems in~\Cite{sun2005convergence}. Furthermore, \Cite{lakshminarayanan2018linear} used a similarity transform based on the solution of Lyapunov equation to analyze stochastic approximation under a constant step-size without specification on the upper bound of the solution of a Lyapunov equation. However, in our case, we can derive the upper bound on the solution of the Lyapunov equation. First, let us consider the following continuous-time Lyapunov equation~\Cite{chen2004linear} for \(T_{Q^*}\): 
\begin{equation}
    GT_{Q^*} + T_{Q^*}^{\top}G = -  I  ,\label{eq:cale}
\end{equation}
where $G = G^{\top}  \succ 0$ is called the Lyapunov matrix. If \(T_{Q^*}\) is Hurwitz, then there exists a Lyapunov matrix \(G\) such that the Lyapunov equation is satisfied, where $G$ is given by 
\begin{equation}
G = \int_0^\infty  {(e^{T_{Q^*} t} )^{\top} (e^{T_{Q^* } t} )dt}. \label{eq:G_solution}
\end{equation}

Before proceeding on, we summarize preliminary results on the properties of $G$. In particular, upper and lower bounds on the spectral norm of \(G\) will play a central role in our analysis.
\begin{lemma}\label{lem:G_upper_bound}\label{lem:inv_G_bound}
\begin{enumerate}[\quad 1)]
    \item $||G||_2$ satisfies 
\begin{align}
\frac{1}{4(|\mathcal{S}|\mathcal{A}|)^{\frac{1}{2}}d_{\max}} \leq ||G||_2 \leq \frac{|\mathcal{S}||\mathcal{A}|}{2(1-\gamma)d_{\min}} \nonumber.
\end{align}
\item  $||G^{-1}||_2$ satisfies
\begin{align}
\frac{2(1-\gamma)d_{\min}}{|\mathcal{S}||\mathcal{A}|} \leq||G^{-1}||_2 \leq  4(|\mathcal{S}||\mathcal{A}|)^{\frac{1}{2}}d_{\max}  \nonumber.
\end{align}
\end{enumerate}
\end{lemma}
The proof is given in Appendix~\Cref{app:lem:G_upper_bound}.
Now, let us consider a similarity transform of \(T_{Q^*}\) using the Lyapunov matrix $G$ as follows:
\begin{equation}\label{eq:B}
    B:=G^{1/2}T_{Q^*}G^{-1/2}.
\end{equation}
Then, we can prove that \(B\) is negative definite, i.e., $\frac{B + B^{\top}}{2} \prec 0$. In particular, it can be proved using the following relations:
\begin{align}\label{eq:lyapunov_eq_B}
    B + B^{\top} &= G^{1/2}T_{Q^*}G^{-1/2} + G^{-1/2}T_{Q^*}^{\top}G^{1/2} = G^{-1/2} (GT_{Q^*} + T^{\top}_{Q^*}G)G^{-1/2} = -  G^{-1} \prec 0. 
\end{align}

In our main development, bounds on $B$ will play an important role as well. 
\begin{lemma}\label{lem:B_upper_bound}
 \(||B||_2\) satisfies
\begin{align}
    \frac{(1-\gamma)d_{\min}}{|\mathcal
    S||\mathcal{A}|} \leq ||B||_2 \leq 2(|\mathcal{S}||\mathcal{A}|)^{\frac{5}{4}} d_{\max} \nonumber.
\end{align}
\end{lemma}
The proof is given in Appendix~\Cref{app:lem:B_upper_bound}. Using the developed similarly transform from $T_{Q^*}$ to $B$ and changing the coordinate of the state in~\eqref{eq:lower_comparison_update}, we can construct another linear system which has more favorable structures. In particular, let us define the new state
\begin{equation}\label{def:z_k}
   z_k := G^{1/2}(Q^L_k-Q).
\end{equation}

Then, the original lower comparison system in~\eqref{eq:lower_comparison_update} can be transformed to
\begin{align*}
        z_{k+1} &= G^{1/2}(Q^L_{k+1}-Q^*) \nonumber\\
            &= G^{1/2}(I+\alpha_k T_{Q^*})(Q^L_k - Q^*)+\alpha_k G^{1/2}w_k \nonumber\\
            &= G^{1/2}(I+\alpha_k T_{Q^*}) G^{-1/2}z_k+\alpha_k G^{1/2}w_k \nonumber\\
            &= (I + \alpha_k B) z_k+\alpha_k G^{1/2} w_k,
\end{align*}
which corresponds to the new state $z_k$. To simplify expressions, let us introduce the notation 
\begin{equation}
    X_k := I + \alpha_k B \label{def:X_k}.
\end{equation}

Using this notation, the last equation can be rewritten as
\begin{align}
z_{k+1} = X_k z_k+\alpha_k G^{1/2} w_k . \label{eq:coordinate_transform_sa}
\end{align}

\begin{figure*}[h!]
\centering
\includegraphics[width=11cm]{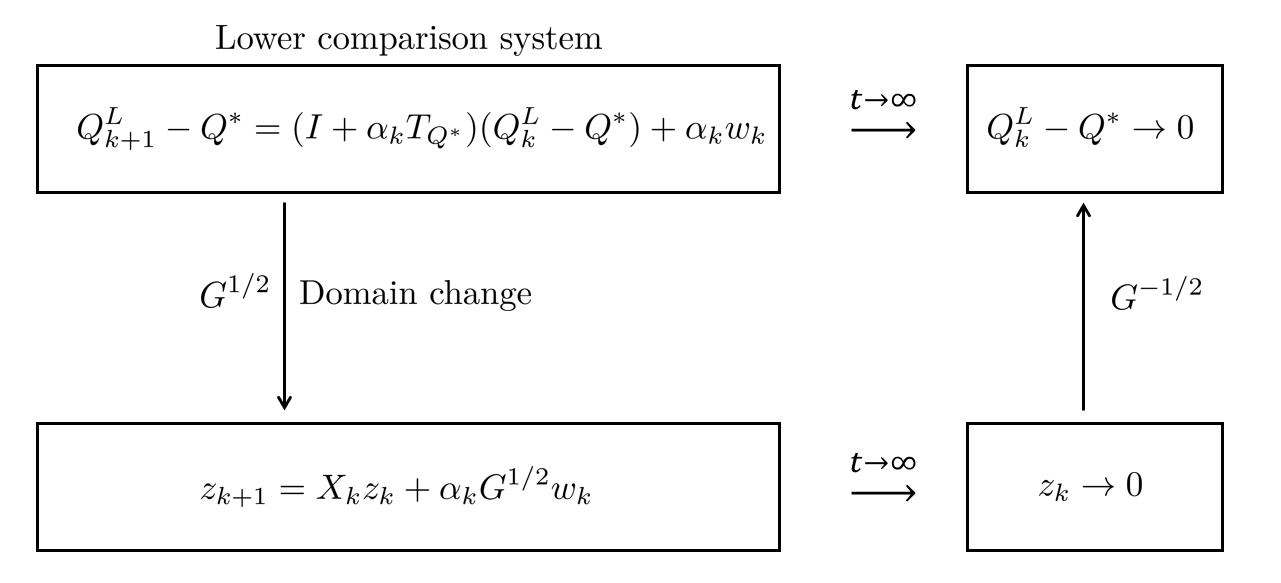}
\caption{Our proof scheme: domain transformation}\label{fig:2}
\end{figure*}

 In this paper, we will analyze the convergence of $z_k$ instead of $Q^L_k-Q^*$. Then, it will ensure the convergence of the original state $Q^L_k-Q^*$. The main benefit of analyzing the system~\eqref{eq:coordinate_transform_sa} comes from negative definiteness of $B$, which will be detailed later. The overall proof scheme is shown in~\Cref{fig:2}.

\subsubsection{Choice of step-size}
In this subsection, we elaborate our motivation for the choice of the step-size \(\alpha_k\). It is known that the diminishing step-size \(\frac{1}{k}\) leads to sample complexity which polynomially depends on \(\frac{1}{1-\gamma}\)~\Cite{even2003learning}. In this paper, we consider the following diminishing step-size:
\begin{align}
    \alpha_k = \frac{\theta}{k+\xi},\label{eq:step_size}
\end{align}
where $\theta >0$ and $\xi > 0$ are tuning parameters to be determined. There are two conditions to be considered in the choice of the tuning parameters, which are summarized below.
\begin{enumerate}
\item The step-size should satisfy the Lyapunov inequality in~\eqref{ineq:dissipativity}.

\item The choice of step-size should lead to (fast) convergence of both lower and upper comparison system.
\end{enumerate} 

In this paper, the two conditions can be satisfied with the following choice of coefficients: 
\begin{enumerate}
\item First of all, $\xi >0$ is selected such that
\begin{align}
      \frac{8\sigma^2_{\max}(B)}{\nu^2}  &\leq \xi  \leq \frac{16\sigma^2_{\max}(B)}{\nu^2} ,  \label{ineq:xi}
\end{align}
where 
\begin{align}\label{def:nu}
\nu := \min \{(1-\gamma)d_{\min} ,\lambda_{\min}(G^{-1})\}.
\end{align}
\item Once $\xi >0$ is chosen based on the above rule, then $\theta$ can be selected as  
\begin{align}
\theta = \frac{\nu}{2\sigma^2_{\max}(B) } \xi.\label{eq:theta}
\end{align}
\end{enumerate}

Note that the upper bound on \( \xi\) is chosen for simplicity of the proof. However, it can be arbitrarily large.
With the above setup, we have the following results.

\begin{lemma}\label{lem:bound on singular value of X_k}
The following statements hold true:
\begin{enumerate}[1)]
    \item Using the step-size in~\eqref{eq:step_size}, we have \begin{align}
         \lambda_{\max}(X^{\top}_k X_k)\leq 1- \beta \alpha_k  ,\label{eq:X_k_bound}
    \end{align}
where 
\begin{align}
\beta:= -\sigma^2_{\max} (B)\alpha_0 + \nu =  \frac{\nu}{2}> 0.\label{eq:beta}
\end{align}

    \item The conditions in~\eqref{ineq:xi} and~\eqref{eq:theta} ensure 
    \begin{align}\label{ineq:tautheta > 2}
    \beta \theta\geq 2,
\end{align}
where $\beta$ is defined in~\eqref{eq:beta}, and
\begin{align}\label{ineq:ctheta>2}
    d_{\min}(1-\gamma)\theta   \geq 2.
\end{align}
\end{enumerate}
\end{lemma}
\begin{proof}
We first prove the first item. The upper bound of \(\lambda_{\max} (I+ \alpha_k (B+B^{\top}) + \alpha_k^2 B^{\top}B) \) can be derived from the relations
\begin{align*}
    \lambda_{\max} (I+ \alpha_k (B+B^{\top}) + \alpha_k^2 B^{\top}B) &= 1 + \alpha_k \lambda_{\max}  ((B+B^{\top}) +\alpha_k B^{\top}B))\\
    &= 1 + \alpha_k \lambda_{\max}  (-G^{-1} +\alpha_k B^{\top}B))\\
    &\leq 1 + \alpha_k (\alpha_k \lambda_{\max}(B^{\top}B) - \lambda_{\min}(G^{-1}))\\
    &= 1 + \alpha_k (\sigma^2_{\max} (B)\alpha_k - \lambda_{\min}(G^{-1})) \\
    &\leq 1 + \alpha_k (\sigma^2_{\max} (B)\alpha_0 - \lambda_{\min}(G^{-1})) \\
    &\leq  1 + \alpha_k (\sigma^2_{\max} (B)\alpha_0 - \nu ),
\end{align*}
where the first inequality follows from the fact that \(G^{-1}\) and \(B^{\top}B\) are positive definite, and the second inequality follows from the fact that \( \alpha_k\) is a non-increasing sequence. The last inequality follows from the definition of \(\nu\) defined in~\eqref{def:nu}. 

Next, we prove the second item. The proof can be readily done by the following simple algebraic inequalities:
\begin{align*}
    \beta \theta \geq  \frac{\nu^2}{4 \sigma^2_{\max}(B) }\xi \geq 2,
\end{align*}
and
\begin{align*}
    d_{\min}(1-\gamma)  \theta \geq d_{\min}(1-\gamma)  \frac{4}{\nu} \geq 2.
\end{align*}
This completes the proof. 
\end{proof}

The first item in~\Cref{lem:bound on singular value of X_k} implies that for the system in the new coordinate~\eqref{eq:coordinate_transform_sa}, the Lyapunov inequality in~\eqref{ineq:dissipativity} is satisfied with \(P=G \) and $p_k = 1 - \beta \alpha_k$ as follows:
\begin{align*}
   A^{\top}_{Q^*,k} G A_{Q^*,k} &= G^{1/2}  G^{-1/2} (I+\alpha_k T_{Q^*})^{\top}G (I+\alpha_k T_{Q^*}) G^{-1/2} G^{1/2} \\
   &= G^{1/2}(I + \alpha_k B)^{\top}(I + \alpha_k B)G^{1/2} \\
   &= G^{1/2}X_k^{\top}X_kG^{1/2} \\
   &\preceq (1-\beta \alpha_k) G,
\end{align*}
where the second equality is due to the definition of \(B\) in~\eqref{eq:B}, and the last inequality follows from the first item of~\Cref{lem:bound on singular value of X_k}. Next, the second item shows that the choice of the diminishing step-size in~\eqref{ineq:xi} and~\eqref{eq:theta} leads to fast convergence of both lower and upper comparison systems. To be more specific, \(\beta\) and \(d_{\min}(1-\gamma)\) are important factors in determining the convergence rate of lower and difference of upper and lower comparison systems, respectively. This can be seen as notions that are analogous to the strong convexity in~\Cite{nemirovski2009robust}. As can been seen in the further analysis, intuitively, each Lyapunov function of the system can be thought of as decreasing by the factors of \( \frac{\beta \theta}{k+\xi} \) and \(\frac{d_{\min}(1-\gamma) \theta}{k+\xi} \) at \(k\)-th iteration. That is, denoting \(V^L_k\) and \(V^U_k\) a the Lyapunov functions of the upper and lower comparison systems at \(k\)-th iteration, we will have
\begin{align*}
    V^L_{k+1} &\leq \left(1-\frac{\beta \theta}{k+\xi} \right)V_k^L + \alpha_kw^L_k,\\
    V^U_{k+1} &\leq \left(1-\frac{d_{\min}(1-\gamma) \theta}{k+\xi} \right)V_k^U+ \alpha_kw^U_k,
\end{align*}
where \(w^L_k\) and \(w^U_k\) are bounded noise term for each system. If \(\theta\) is small, the decreasing rate would be small, leading to slow convergence. Therefore, the step-size should be chosen to reflect these properties.  



\begin{remark}
Note that our choice of step-size can be expressed in terms of \( d_{\min}, |\cal{S}||\cal{A}|,\) and \(1-\gamma\), since we can bound \( \sigma^2_{\max} (B)\) from~\Cref{lem:B_upper_bound}. However, in the sequel, we keep using \(\sigma^2_{\max} (B) \) in the choice of step-sizes for convenience and simplicity of the proof.
\end{remark}


\subsubsection{Proof of convergence rate for lower comparison system}
We now prove the convergence rate for~\eqref{eq:lower_comparison_update}, beginning with convergence rate for \(z_k\). To this end, we will consider the simple quadratic Lyapunov function candidate 
\begin{align*}
V(z) = z^{\top} z,\quad z\in {\mathbb R}^{|{\cal S}||{\cal A}|}.
\end{align*}
Convergence of the Lyapunov function candidate is established in the sequel. 

\begin{proposition}\label{prop:lower_z_convergence_rate}
Consider the Lyapunov function candidate \( V(z) = z^{\top} z \) in~\cref{eq:coordinate_transform_sa}. Under the step-size condition in~\eqref{eq:step_size}, we have
\begin{align}
    \mathbb{E}[V(z_k)] \leq \frac{1}{k+ \xi} \max \left\{ \frac{8(|\mathcal{S}||\mathcal{A}|)^{\frac{11}{2}}d_{\max}^2}{(1-\gamma)^3d_{\min}^3}||Q^L_0 -Q^*||^2_2 ,
   \frac{72(|\mathcal{S}||\mathcal{A}|)^3}{(1-\gamma)^5d_{\min}^3}
    \right\} \nonumber.
\end{align}
\end{proposition}
\begin{proof}
We first derive an upper bound on \(\mathbb{E}[V(z_{k+1})|\mathcal{F}^L_k]\) from the relations
\begin{align}
    \mathbb{E}[V(z_{k+1})|\mathcal{F}^L_k] &= \mathbb{E}[(z_k)^{\top}X_k^{\top}X_k(z_k)+ 2\alpha_k z_k^{\top}X_k^{\top}G^{1/2}w_k + \alpha_k^2 w_k^{\top}Gw_k| \mathcal{F}^L_k]   \label{eq:v_low_expand} \\
    &\leq (z_k)^{\top}X_k^{\top}X_k(z_k) +\alpha_k^2\lambda_{\max}(G)\frac{9}{(1-\gamma)^2} \nonumber \\
                                        &\leq (1-\beta \alpha_k ) V(z_k)  + \alpha_k^2\lambda_{\max}(G)\frac{9}{(1-\gamma)^2} \nonumber \\
                                        &= \left(1- \frac{\beta\theta}{k+\xi}\right)V (z_k) + 
                                        \frac{\theta^2}{(k+\xi)^2}\lambda_{\max}(G)\frac{9}{(1-\gamma)^2}\nonumber\\
                                        &\leq \left(1- \frac{2}{k+\xi}\right)V (z_k) + 
                                        \frac{1}{(k+\xi)^2}  \frac{16(|\mathcal{S}||\mathcal{A}|)^2}{(1-\gamma)^2d_{\min}^2} \frac{|\mathcal{S}||\mathcal{A}|}{2(1-\gamma)d_{\min}}
                                          \frac{9}{(1-\gamma)^2} \nonumber \\
                                        &\leq \left(1- \frac{2}{k+\xi}\right)V (z_k) + 
                                        \frac{1}{(k+\xi)^2}   \frac{72(|\mathcal{S}||\mathcal{A}|)^3}{(1-\gamma)^5d_{\min}^3},\nonumber
\end{align}
where \(\mathcal{F}^L_k\) is defined as the \(\sigma\)-field induced by \(\{ (s_i,a_i,s_{i}',r_i, Q_i,Q^L_i)\}_{i=0}^k\). Here, the first inequality follows from
\begin{equation*}
    \mathbb{E}[2\alpha_k z_k^{\top}X_k^{\top}G^{1/2}w_k|\mathcal{F}^L_k] = 0,
\end{equation*}
and the bound on \(w_k\) given in~\eqref{ineq:bound on w_k}. The second inequality comes from the bound given in~\Cref{lem:bound on singular value of X_k}. The third inequality is due to~\eqref{ineq:tautheta > 2}, and the last inequality is obtained simply by collecting the terms in the previous equations. Now, taking the total expectation, one gets
\begin{align}
     \mathbb{E}[V(z_{k+1})] \leq \left(1- \frac{2}{k+\xi}\right)\mathbb{E}[V (z_k)] + 
                                        \frac{1}{(k+\xi)^2}\frac{72(|\mathcal{S}||\mathcal{A}|)^3}{(1-\gamma)^5d_{\min}^3} \nonumber.
\end{align}

To proceed further, an induction argument is applied. For simplicity, let us define the notation \( \hat{k}:= k +\xi \), and suppose that
\begin{align}
\mathbb{E}[V(z_{k})] \leq \frac{1}{\hat{k}}\max\left\{\xi||z_0||^2_2,\frac{72(|\mathcal{S}||\mathcal{A}|)^3}{(1-\gamma)^5d_{\min}^3}\right\} \label{ineq:z_induc_k} 
\end{align}
holds. We now prove that \( \mathbb{E}[V(z_{k+1})]  \leq \frac{1}{\hat{k}+1} \max\left\{\xi||z_0||^2_2,\frac{72(|\mathcal{S}||\mathcal{A}|)^3}{(1-\gamma)^5d_{\min}^3}\right\}\) holds. It is done by the inequalities
\begin{align}
     \mathbb{E}[V(z_{k+1})] &\leq  \left(1-\frac{2}{\hat{k}}\right)\frac{1}{\hat{k}} + \frac{1}{\hat{k}^2} \max\left\{\xi||z_0||^2_2,\frac{72(|\mathcal{S}||\mathcal{A}|)^3}{(1-\gamma)^5d_{\min}^3}\right\}\nonumber\\
     &\leq \left(\frac{1}{\hat{k}}  - \frac{1}{\hat{k}^2}\right) \max\left\{\xi||z_0||^2_2,\frac{72(|\mathcal{S}||\mathcal{A}|)^3}{(1-\gamma)^5d_{\min}^3}\right\}\nonumber\\
     &\leq \frac{1}{\hat{k}+1} \max\left\{\xi||z_0||^2_2,\frac{72(|\mathcal{S}||\mathcal{A}|)^3}{(1-\gamma)^5d_{\min}^3}\right\} \nonumber.
\end{align} 
Hence, the inequality at \(k+1\) is satisfied.
Next, using~\Cref{lem:xi_bound} given in Appendix~\Cref{app:sec:step_size}, we can bound \(\xi ||z_0||_2^2\) as follows:
\begin{align}
    \xi ||z_0||^2_2 &\leq \frac{16(|\mathcal{S}||\mathcal{A}|)^{\frac{9}{2}}d_{\max}^2}{(1-\gamma)^2d_{\min}^2}|| G||_2 ||Q^L_0 -Q^*||^2_2 \nonumber\\
                    &\leq \frac{16(|\mathcal{S}||\mathcal{A}|)^{\frac{9}{2}}d_{\max}^2}{(1-\gamma)^2d_{\min}^2} \frac{|\mathcal{S}||\mathcal{A}|}{2(1-\gamma)d_{\min}} ||Q^L_0 -Q^*||^2_2 \nonumber\\
                    &\leq  \frac{8(|\mathcal{S}||\mathcal{A}|)^{\frac{11}{2}}d_{\max}^2}{(1-\gamma)^3d_{\min}^3}||Q^L_0 -Q^*||^2_2 \nonumber.
\end{align}
Applying above inequality to~\eqref{ineq:z_induc_k} and by the induction argument, we have
\begin{align}
    \mathbb{E}[V(z_{k})] \leq \frac{1}{k+ \xi} \max \left\{ \frac{8(|\mathcal{S}||\mathcal{A}|)^{\frac{11}{2}}d_{\max}^2}{(1-\gamma)^3d_{\min}^3}||Q^L_0 -Q^*||^2_2 ,
   \frac{72(|\mathcal{S}||\mathcal{A}|)^3}{(1-\gamma)^5d_{\min}^3}
    \right\} \nonumber,
\end{align}
which is the desired conclusion.
\end{proof}

\Cref{prop:lower_z_convergence_rate} provides \(\mathcal{O}(\frac{1}{k})\) convergence rate for ${\mathbb E}[V(z_k )] = {\mathbb E}[\left\| {z_k } \right\|_2^2 ]$ instead of ${\mathbb E}[V(Q_k^L-Q^*)] = {\mathbb E}[\left\| Q_k^L-Q^* \right\|_2^2 ]$. Since \(Q_k^L-Q^* = G^{-1/2}(z_k)\), it directly leads to the convergence rate corresponding to \(Q_k^L-Q^*\), which can be derived with simple algebraic inequalities. The following theorem establishes its convergence rate.
\begin{theorem}\label{thm:lower_system_convergence_rate}
Under the step-size~\eqref{eq:step_size}, we have
\begin{align}
    \mathbb{E}[||Q^L_k-Q^*||_{\infty}] \leq \frac{1}{\sqrt{k+\xi}} \max \left\{ \frac{4(|\mathcal{S}||\mathcal{A}|)^3d_{\max}^{\frac{3}{2}}}{(1-\gamma)^{\frac{3}{2}}d_{\min}^{\frac{3}{2}}}||Q^L_0 -Q^*||_2 ,
    \frac{14 (|\mathcal{S}| |\mathcal{A}|)^{\frac{7}{4}}d_{\max}^{\frac{1}{2}}}{(1-\gamma)^{\frac{5}{2}}d_{\min}^{\frac{3}{2}}}     \right\} \nonumber.
\end{align}
\end{theorem}
\begin{proof}
Since \(Q^L_k-Q^* = G^{-1/2}(z_k) \), we have
\begin{align}
       \mathbb{E}[||Q^L_k-Q^*||^2_2] &= \mathbb{E}[z^{\top}_k G^{-1} z_k] \nonumber\\
        &\leq \lambda_{\max}(G^{-1}) \mathbb{E}[||z_k||^2_2] \nonumber\\
        &\leq  \frac{4(|\mathcal{S}||\mathcal{A}|)^{\frac{1}{2}}d_{\max} }{k+ \xi} \max \left\{ \frac{2(|\mathcal{S}||\mathcal{A}|)^{\frac{11}{2}}d_{\max}^2}{(1-\gamma)^3d_{\min}^3}||Q^L_0 -Q^*||^2_2 ,
   \frac{72(|\mathcal{S}||\mathcal{A}|)^3}{(1-\gamma)^5d_{\min}^3}
    \right\} \nonumber \\
    &=\frac{1}{k+ \xi} \max \left\{\frac{8(|\mathcal{S}||\mathcal{A}|)^6 d_{\max}^3}{(1-\gamma)^3 d_{\min}^3}||Q^L_0 -Q^*||^2_2 ,
    \frac{288 (|\mathcal{S}| |\mathcal{A}|)^{\frac{7}{2}}d_{\max}}{(1-\gamma)^{5}d_{\min}^3}
    \right\} \nonumber,
\end{align}
where the first inequality is due to the positive definiteness of \(G^{-1}\), and the second inequality follows from~\Cref{prop:lower_z_convergence_rate}. Applying Jensen's inequality to the above result yields
\begin{align*}
    \mathbb{E}[||Q^L_k-Q^*||_{\infty}] \leq& \mathbb{E}[||Q^L_k-Q^*||_2]\\
    \leq& \sqrt{\mathbb{E}[||Q^L_k-Q^*||^2_2]}\\
    \leq& \frac{1}{\sqrt{k+\xi}} \max \left\{ \frac{4(|\mathcal{S}||\mathcal{A}|)^3d_{\max}^{\frac{3}{2}}}{(1-\gamma)^{\frac{3}{2}}d_{\min}^{\frac{3}{2}}}||Q^L_0 -Q^*||_2 ,
    \frac{14 (|\mathcal{S}| |\mathcal{A}|)^{\frac{7}{4}}d_{\max}^{\frac{1}{2}}}{(1-\gamma)^{\frac{5}{2}}d_{\min}^{\frac{3}{2}}}     \right\},
\end{align*}
which is the desired conclusion.
\end{proof}

The convergence rate of lower comparison system depends on  \( (1-\gamma)^{-\frac{5}{2}}d_{\min}^{-\frac{3}{2}} \). As can be seen in the sequel, the upper comparison system has slower convergence rate then lower comparison system. The bottleneck of convergence rate of Q-learning may be due to the relatively slower convergence of the upper comparison system.

%% file: upper_system.tex
In this section, we focus on the construction of the upper comparison system. Then, its convergence analysis is discussed.

\subsubsection{Construction of upper comparison system}
As in~\Cite{lee2021discrete}, the upper comparison system in~\eqref{eq:simplified_qlearning_sa} can be derived as follows:
\begin{align}~\label{eq:upper}
    Q^U_{k+1} - Q^* = A_{Q_k,k} (Q_k^U-Q^* )+ \alpha_k w_k.
\end{align}
For completeness, we provide the proof that \(Q^U_k\) upper bounds \(Q_k\). 
\begin{lemma}[\Cite{lee2021discrete}]
If \( Q^U_0 - Q^* \geq Q_0  - Q^*\), then
\begin{align}
Q^U_k - Q^* \geq Q_k - Q^* \nonumber
\end{align}
for all \(k \geq 0 \).
\end{lemma}
\begin{proof}
The proof follows from induction.
To this end, suppose that \( Q^U_{k} - Q^* \geq Q_k - Q^*\) holds for some \(k\geq 0\). Then, we have
\begin{align}
    Q^U_{k+1} - Q^*  &=  A_{Q_k,k} (Q_k^U-Q^* )+ \alpha_k w_k  \nonumber \\
                    &\geq  A_{Q_k,k} (Q_k^U-Q^* )+ b_{Q_k,k}+\alpha_k w_k \nonumber \\
                    &\geq A_{Q_k,k} (Q_k-Q^* )+ b_{Q_k,k}+\alpha_k w_k \nonumber\\
                    &= Q_{k+1} - Q^* \nonumber,
\end{align}
where the first inequality follows from the fact that \(b_{Q_k,k} \leq 0 \). The second inequality is due to positiveness of \(A_{Q_k,k}\). This completes the proof.
\end{proof}

Since the upper comparison system in~\eqref{eq:upper} is a switching linear system, it is not possible to obtain the corresponding convergence rate with the same techniques as in the lower comparison system. Moreover, we cannot find a common similarity transform matrix as in the lower comparison system case. Instead, we analyze the difference between \(Q^U_k\) and \(Q^L_k\) as in~\Cite{lee2021discrete}. In particular, we consider the difference system
\begin{align}
    Q^U_{k+1} - Q^L_{k+1} &= A_{Q_k,k} (Q_k^U-Q^k_L) + B_{Q_k,k}(Q^L_k-Q^*)   \label{eq:up_lr_diff}
\end{align}
where
\begin{align}
    B_{Q_k,k} &:= A_{Q_k,k} - A_{Q^*,k} = \alpha_k \gamma DP (\Pi_{Q_k}-\Pi_{Q^*}) \nonumber.
\end{align} 
\subsubsection{Proof of convergence rate for difference of upper and lower comparison systems}

With the above setup, we are now ready to give the convergence rate of \( \mathbb{E} [|| Q^U_{k}- Q^L_{k} ||_{\infty}] \). For simplicity of the proof, we assume \( Q^L_0 = Q^U_0 \) without loss of generality.
\begin{theorem}\label{thm:upper_system_convergence_rate}
Under the diminishing step-size in~\eqref{eq:step_size}, we have
\begin{align}
      \mathbb{E} [|| Q^U_{k}- Q^L_{k} ||_{\infty}] &\leq 
      \frac{1}{\sqrt{k+\xi}} \max \left\{\frac{16(|\mathcal{S}||\mathcal{A}|)^4d_{\max}^{\frac{5}{2}}
      }{(1-\gamma)^{\frac{5}{2}}d_{\min}^{\frac{5}{2}}}||Q^L_0 -Q^*||_2 ,
    \frac{56 (|\mathcal{S}| |\mathcal{A}|)^{\frac{11}{4}} d_{\max}^{\frac{3}{2}}}{(1-\gamma)^{\frac{7}{2}}d_{\min}^{\frac{5}{2}}} \right\} .
\end{align}
\end{theorem}
\begin{proof}
The proof follows similar lines as~\Cref{thm:lower_system_convergence_rate}. We first take an expectation on the difference system~\eqref{eq:up_lr_diff}, and use sub-multiplicativity of the norm as follows:
\begin{align}
   \mathbb{E} [ || Q^U_{k+1}- Q^L_{k+1} ||_{\infty}] &\leq \mathbb{E} [||A_{Q_k,k}||_{\infty} ||Q^U_k - Q^L_k||_{\infty} ]+ \mathbb{E} [||B_{Q_k}||_{\infty} ||Q^L_k - Q^*||_{\infty}]  \nonumber \\
    &\leq \left( 1- (1-\gamma)d_{\min}\alpha_k\right) \mathbb{E} [||Q^U_k - Q^L_k ||_{\infty}] \\
    &+ 2\alpha_k \gamma d_{\max} \mathbb{E} [||Q^L_k - Q^*||_{\infty}] \nonumber \\
    &\leq \left( 1- \frac{2}{k+\xi} \right) \mathbb{E} [||Q^U_k - Q^L_k ||_{\infty}] +\frac{2\theta \gamma d_{\max}}{k+\xi} \frac{  C_L}{\sqrt{k+\xi}}  \nonumber \\
    &\leq \left( 1- \frac{2}{k+\xi} \right) \mathbb{E} [||Q^U_k - Q^L_k ||_{\infty}]+ \frac{C_U}{(k+\xi)\sqrt{k+\xi}}  \nonumber
\end{align}
where
\begin{align*}
    C_L&:= \max \left\{ \frac{4(|\mathcal{S}||\mathcal{A}|)^3d_{\max}^{\frac{3}{2}}}{(1-\gamma)^{\frac{3}{2}}d_{\min}^{\frac{3}{2}}}||Q^L_0 -Q^*||_2 ,
    \frac{14 (|\mathcal{S}| |\mathcal{A}|)^{\frac{7}{4}}d_{\max}^{\frac{1}{2}}}{(1-\gamma)^{\frac{5}{2}}d_{\min}^{\frac{3}{2}}}     \right\},\\
        C_U&:= \max \left\{\frac{16(|\mathcal{S}||\mathcal{A}|)^4 d_{\max}^{\frac{5}{2}}}{(1-\gamma)^{\frac{5}{2}}d_{\min}^{\frac{5}{2}}}||Q^L_0 -Q^*||_2 ,
    \frac{56 (|\mathcal{S}| |\mathcal{A}|)^{\frac{11}{4}} d_{\max}^{\frac{3}{2}}}{(1-\gamma)^{\frac{7}{2}}d_{\min}^{\frac{5}{2}}} \right\}.
\end{align*} 
The second inequality follows from~\Cref{lem:spectral_radius_of_A}, the third inequality is due to~\eqref{ineq:ctheta>2}, and the last inequality comes from~\Cref{lem:theta_bound} given in Appendix~\Cref{app:sec:sample_complexity}. We proceed by induction. Let \( \hat{k} = k + \xi\), and suppose \( \mathbb{E} [ ||Q^U_k - Q^L_k||_{\infty}] \leq \frac{C_U}{\sqrt{\hat{k}}} \) for some \(k\geq0\).
Now, we prove that \(\mathbb{E} [ ||Q^U_{k+1} - Q^L_{k+1}||_{\infty}] \leq \frac{C_U}{\sqrt{\hat{k}+1}}\) holds as follows:
\begin{align}
     \mathbb{E} [|| Q^U_{k+1}- Q^L_{k+1} ||_{\infty}] &\leq  \frac{C_U}{\sqrt{\hat{k}}} - \frac{2C_U}{\hat{k}\sqrt{\hat{k}}} + \frac{C_U}{\hat{k}\sqrt{\hat{k}}} \nonumber\\
                                         &\leq \frac{C_U}{\sqrt{\hat{k}}} - \frac{C_U}{\hat{k}\sqrt{\hat{k}}} \nonumber\\
                                         &\leq \frac{C_U}{\sqrt{\hat{k}+1}} \nonumber.
\end{align}
To prove the last inequality, we need to show \( \frac{1}{\sqrt{\hat{k}}} - \frac{1}{\hat{k}\sqrt{\hat{k}}} \leq \frac{1}{\sqrt{\hat{k}+1}}\).
Multiplying \(\hat{k}^{\frac{3}{2}}\sqrt{\hat{k}+1}\) on both sides and rearranging the terms lead to \(  (\hat{k}-1) \sqrt{\hat{k}+1} \leq \hat{k}\sqrt{\hat{k}}\). Now, it suffices to prove the last inequality. If \( 0< \hat{k} < 1\), the inequality is trivial.
When \( \hat{k} \geq 1\), taking square on both sides, we have
\begin{align}
    (\hat{k}^2-2\hat{k}+1)(\hat{k}+1) &\leq \hat{k}^3 \nonumber.
\end{align}
Rearranging the terms, the inequality becomes
\begin{align}
    -2\hat{k}^2+\hat{k} + \hat{k}^2-2\hat{k}+1 &= -\hat{k}^2 -\hat{k} +1 \leq 0 , \nonumber
\end{align}
which proves the inequality \(  (\hat{k}-1) \sqrt{\hat{k}+1} \leq \hat{k}\sqrt{\hat{k}}\). This completes the proof.
\end{proof}

Compared to the convergence rate of the lower comparison system given in~\eqref{eq:lower_comparison_update}, the convergence rate for \(Q^U_k-Q^L_k\) given in~\eqref{eq:up_lr_diff} depends on \((1-\gamma)^{-\frac{7}{2}}d_{\min}^{-\frac{5}{2}}\), which leads to slower convergence of the system.



%% file: original_system.tex
With the bounds on upper and lower comparison systems, we are now ready to bound the original system in this section. We use simple algebraic inequalities to bound the original iterate \({\mathbb E}[||Q
_k-Q^*||_{\infty}]\).
\begin{theorem}\label{thm:iid_convergence_rate}
Under the step-size in~\eqref{eq:step_size}, we have
\begin{align}
  \mathbb{E} [  ||Q_k - Q^*||_{\infty}] \leq \frac{1}{\sqrt{k+\xi}} \max \left\{\frac{32(|\mathcal{S}||\mathcal{A}|)^4d_{\max}^{\frac{5}{2}}}{(1-\gamma)^{\frac{5}{2}}d_{\min}^{\frac{5}{2}}}||Q^L_0 -Q^*||_2 ,
    \frac{112 (|\mathcal{S}| |\mathcal{A}|)^{\frac{11}{4}} d_{\max}^{\frac{3}{2}}}{(1-\gamma)^{\frac{7}{2}}d_{\min}^{\frac{5}{2}}} \right\}  \nonumber.
\end{align}
\end{theorem}
\begin{proof}
The inequality follows from applying triangle inequality of the norm as follows:
\begin{align}
     \mathbb{E} [||Q_k - Q^*||_{\infty}] &\leq  \mathbb{E} [||Q^*-Q^L_k||_{\infty}] + \mathbb{E} [||Q_k - Q^L_k||_{\infty} ]\nonumber \\
    &\leq \mathbb{E} [||Q^*-Q^L_k||_{\infty}] + \mathbb{E} [||Q^U_k-Q^L_k||_{\infty}] \nonumber \\
    &\leq \frac{1}{\sqrt{k+\xi}} \max \left\{ \frac{4(|\mathcal{S}||\mathcal{A}|)^3d_{\max}^{\frac{3}{2}}}{(1-\gamma)^{\frac{3}{2}}d_{\min}^{\frac{3}{2}}}||Q^L_0 -Q^*||_2 ,
    \frac{14 (|\mathcal{S}| |\mathcal{A}|)^{\frac{7}{4}}d_{\max}^{\frac{1}{2}}}{(1-\gamma)^{\frac{5}{2}}d_{\min}^{\frac{3}{2}}}     \right\} \nonumber\\
    &+\frac{1}{\sqrt{k+\xi}} \max \left\{\frac{16(|\mathcal{S}||\mathcal{A}|)^4d_{\max}^{\frac{5}{2}}}{(1-\gamma)^{\frac{5}{2}}d_{\min}^{\frac{5}{2}}}||Q^L_0 -Q^*||_2 ,
    \frac{56 (|\mathcal{S}| |\mathcal{A}|)^{\frac{11}{4}} d_{\max}^{\frac{3}{2}}}{(1-\gamma)^{\frac{7}{2}}d_{\min}^{\frac{5}{2}}} \right\} \nonumber \\
    &\leq \frac{1}{\sqrt{k+\xi}} \max \left\{\frac{32(|\mathcal{S}||\mathcal{A}|)^4d_{\max}^{\frac{5}{2}}}{(1-\gamma)^{\frac{5}{2}}d_{\min}^{\frac{5}{2}}}||Q^L_0 -Q^*||_2 ,
    \frac{112 (|\mathcal{S}| |\mathcal{A}|)^{\frac{11}{4}} d_{\max}^{\frac{3}{2}}}{(1-\gamma)^{\frac{7}{2}}d_{\min}^{\frac{5}{2}}} \right\} \nonumber,
\end{align}
where the first inequality follows from triangle inequality. The second inequality is due to the fact that \(||Q_k-Q^L_k||_{\infty} \leq ||Q^U_k-Q^L_k ||_{\infty}\). This is because $0 \le Q_k  - Q_k^L  \le Q_k^U  - Q_k^L$ holds. Moreover, the third inequality comes from~\Cref{thm:lower_system_convergence_rate} and~\Cref{thm:upper_system_convergence_rate}.
This completes the proof.
\end{proof}

Since \(\mathbb{E}[||Q^U_k - Q^L_k]||_{\infty}]\) has a worse bound than \(\mathbb{E}[||Q^L_k - Q^*||_{\infty}]\), convergence of \(\mathbb{E}[||Q_k-Q^*||_{\infty}]\) is dominated by the bounds on \(\mathbb{E}[||Q^U_k - Q^L_k]||_{\infty}]\).




%% file: markovian/markovian_copy.tex
In the previous section, our analysis was based on i.i.d. observation model. However, in real-world scenarios, the state-action tuple follows Markovian observation model. In this section, we assume that the state-action pair follows a Markov chain as summarized in~\Cref{sec:revisit-Q-learning}. The corresponding Q-learning algorithm with Markovian observation model in~\cref{algo:standard-Q-learning2} can be written as follows:
\begin{equation}\label{eq:q_learning_sa}
    Q_{k+1} = Q_k +\alpha_k (D_\infty R + \gamma D_\infty P\Pi_{Q_k}-D_\infty Q_k + w_k) ,
\end{equation}
where $D_\infty  \in {\mathbb R}^{|{\cal S}||{\cal A}| \times |{\cal S}||{\cal A}|}$ is a diagonal matrix with diagonal elements being the stationary state-action distributions, i.e., 
\begin{align*}
(e_s  \times e_a )^{\top} D_\infty  (e_s  \times e_a ) = \mu _\infty  (s,a) = \lim_{k \to \infty } {\mathbb P}[s_k  = s,a_k  = a|\beta ],
\end{align*}
and
\begin{align}
    w_k &:= (e_{a_k} \otimes e_{s_k} ) (e_{a_k} \otimes e_{s_k})^{\top} R + \gamma (e_{a_k} \otimes e_{s_k})(e_{s_{k+1}})^{\top}\Pi_{Q_k}Q_k \label{def:w_k_markov} \\
    &- (e_{a_k} \otimes e_{s_k})(e_{a_k}\otimes e_{s_k})^{\top} Q_k \nonumber\\
    &- (D_\infty R+ \gamma D_\infty P\Pi_{Q_k}Q_k - D_\infty Q_k) \nonumber
\end{align}
is a noise term. 
Using the Bellman equation in~\eqref{eq:bellman_optimal_q}, one can readily prove that the error term \(Q_k - Q^* \) evolves according to the following error dynamic system:
\begin{align*}
 Q_{k+1}-Q^* =    A_{Q_k,k}(Q_k - Q^*)+b_{Q_k,k}+\alpha_k w_k,
\end{align*}
where
\begin{align}\label{def:A,b,markov}
    A_{Q,k} &:= I+\alpha_k T_{Q} \quad b_{Q,k}:= \alpha_k  \gamma D_\infty P(\Pi_Q - \Pi_{Q^*})Q^* \\
    T_{Q} &:= \gamma D_\infty P\Pi_{Q}-D_\infty \nonumber.
\end{align}

Note that we can view \(A_{Q,k}\) as a system matrix of a stochastic linear time-varying switching system as in the previous section. With slight abuse of notation, we use  the notation \(G\) for the solution of Lyapunov equation~\eqref{eq:cale} with \(T_{Q^*}\) in~\eqref{def:A,b,markov}, and \(B:=G^{1/2}T_{Q^*}G^{-1/2}\), the similarity transform of \(T_{Q^*}\) to a negative definite matrix, and \(X_k := I+\alpha_k B\). \(T_{Q^*}\) in~\eqref{def:A,b,markov} can be thought of as substituting \(D\) in~\eqref{def:A,b} with \(D_{\infty}\). Since \(G\), \(B\), and \(X_k\) only depend on \(T_{Q^*}\), related lemmas still hold with \(d_{\min}\) and \(d_{\max}\) defined in~\Cref{sec:revisit-Q-learning} for the Markovian case.  Moreover, due to the correlation of \(Q_k\) and \((s_k,a_k,s_{k+1},r_k)\), the noise term is biased, i.e., \( \mathbb{E}[w_k|\mathcal{F}_k]\neq 0 \), where \(\mathcal{F}_k\) is defined as the \(\sigma\)-field induced by \(\{ (s_i,a_i,s_{i+1},r_i, Q_i)\}_{i=0}^k\). Therefore, the so-called crossing term in~\eqref{eq:v_low_expand}
\begin{align}\label{neq:gradient_bias}
   \xi_k(z_k,Q_k) :=  \xi(z_k,Q_k,o_k) := z^{\top}_kX_k^{\top}G^{1/2}w_k(Q_k).
\end{align} 
is not zero, where \( o_k := (s_k,a_k,r_{k+1},s_{k+1}) \) is the observation, which is the main difference compared to the analysis in the previous section. To resolve the difficulty, in this paper, we assume that the distribution $\mu_k$ mixes in a geometric rate to the stationary distribution \(\mu_\infty\), i.e., $\mu_k \to \mu_\infty$ as $k \to \infty$ exponentially. Note that this assumption is standard and common in the literature~\Cite{sun2018markov,bhandari2018finite,duchi2012ergodic,chen2021lyapunov,li2020sample}, which study stochastic optimization algorithms with Markovian observation models. With the help of geometric mixing rate of the underlying Markov chain, our aim is to bound~\eqref{neq:gradient_bias} by \( \mathcal{O}\left(\frac{\tau^{\mix}(\alpha_k)}{k}\right) \). 
The previous study in~\Cite{bhandari2018finite} has proved that such biased noise in standard TD-learning can be bounded by \( \mathcal{O}\left(\frac{\tau^{\mix}(\alpha_k)}{k}\right) \) with an additional projection step. In this paper, we prove that with our setups, we can easily follow the spirits of~\Cite{bhandari2018finite}, and prove convergence of Q-learning without the additional projection step.

Finally, in this section, we separate the index of the observation \(\{o_k\}_{k=1}^{n}\)and the iterate \( \{Q_k \}_{k=1}^{n} \) in the noise term, i.e.,
\begin{align*}
    w_j(Q_i) &:= (e_{a_j} \otimes e_{s_j} ) (e_{a_j} \otimes e_{s_j})^{\top} R + \gamma (e_{a_j} \otimes e_{s_j})(e_{s_{j+1}})^{\top}\Pi_{Q_i}Q_i\\
    &- (e_{a_j} \otimes e_{s_j})(e_{a_j}\otimes e_{s_j})^{\top} Q_i \\
    &- (D_\infty R+ \gamma D_\infty P\Pi_{Q_i}Q_i - D_\infty Q_i).
\end{align*}
Moreover, we will use the notation \(w(o_j,Q_i):=w_j(Q_i)\) interchangeably in order to specify the dependence of the noise on the observation. The subscript of \(w_j\) indicates the time index of observation, while the subscript of $Q_i$ is the time index corresponding to Q-iteration. When their indices coincide, that is, having both  \(k\)-th observation of \(\{o_k\}_{k=1}^{n}\) and \(k\)-the iterate of \( \{Q_k \}_{k=1}^{n} \), we use \(w_k\). Similar to the i.i.d. case, we can bound the noise term as follows:

\begin{lemma}\label{lem:markovian_noise_bound}
Under Markovian observation model in~\eqref{def:w_k_markov}, the stochastic noise can be bounded as follows:
\begin{align*}
    ||w_k||_{\infty} &\leq \frac{4}{1-\gamma}  ,\\
    \mathbb{E}[||w_k||^2_2 \mid \mathcal{F}_k] &\leq \frac{16|\mathcal{S}||\mathcal{A}|}{(1-\gamma)^2},
\end{align*}
where \( \mathcal{F}_k\) is the \(\sigma\)-field induced by \(\{ (s_i,a_i,s_{i+1},r_i, Q_i)\}_{i=0}^k\).
\end{lemma}
The proof is in Appendix~\Cref{app:lem:markovian_noise_bound}.

\subsection{Preliminaries on information theoretic results}\hfill\\
In this subsection, we introduce some definitions used in information theory, and its relation to the Markov chain under our consideration.
\begin{definition}[Total variation distance~\Cite{levin2017markov}]\label{def:tv}
The total variation distance between two probability distributions \(P\) and \(Q\) on the sample space, \(\Omega\), is defined as
\begin{align*}
    d_{\tv}(P,Q) := \max_{A\in \Omega} | P(A) - Q(A)|.
\end{align*}
\end{definition}

\begin{definition}[Conditional independence~\Cite{koller2009probabilistic}]\label{def:conditional-independence}
For random variables X,Y, and Z, 
\begin{align*}
    X \to Y \to Z
\end{align*}
denotes that \(X\) and \(Z\) are conditionally independent given \(Y\), that is, ${\mathbb P}(Z|X,Y) = {\mathbb P}(Z|Y)$.
\end{definition}
Throughout the analysis, we will consider standard assumptions on Markov chain. With~\Cref{assmp:irreducible_aperiodic}, the Markov chain reaches its stationary distribution at a geometric rate. We note that most of the analysis of stochastic optimization on Markov chain~\Cite{sun2018markov,bhandari2018finite,duchi2012ergodic,chen2021lyapunov,li2020sample} relies on geometric convergence to the stationary distribution. Our analysis also adopts this standard scenario. 
\begin{lemma}[\Cite{levin2017markov}]\label{assumption:mixing}
Under~\Cref{assmp:irreducible_aperiodic}, there exist constants \(m>0\) and \( \rho \in (0,1) \) such that 
\begin{equation*}
     d_{\tv}(\mu_k,\mu_\infty) \leq m \rho^k, \quad \forall{k} \in \mathbb{N},
\end{equation*}
where $\mu_k$ and $\mu_\infty$ are defined in~\eqref{eq:mu_k_infty}.
\end{lemma}

Now, we can define the mixing-time as follows:
\begin{align}\label{def:mixing_time}
    \tau^{\mix}(c) := \min \{k\in \mathbb{N} \mid m \rho^k \leq c\},
\end{align}
which is the minimum time required for the total variation distance between the stationary distribution and the current distribution to be less than a desired level $c >0$. Moreover, \(m\) and \(\rho\) are the constants defined in~\Cref{assumption:mixing}.

Related to the mixing-time, let us introduce the following quantity:
\begin{align}
    K_{\mix} := \min \{ k \mid  t \geq  2\tau^{\mix}(\alpha_t) , \forall{t\geq k}\} \label{def:K},
\end{align}
where \(\alpha_t\) is the step-size defined in~\eqref{eq:step_size}. Note that \( K_{\mix}\) is well-defined, since upper bound of \( \tau^{\mix} (\alpha_k)\) is logarithmically proportional to \(k\). Intuitively, \( K_{\mix}\) roughly means the number of iterations, after which the mixing-time corresponding to the current step-size can be upper bounded by the current time. Therefore, the analysis can be divided into two phases, the first phase $k \le K_{\mix}$ and the second phase $k \ge K_{\mix}$. Note that this approach is common in the literature. For example, the recent paper~\Cite{chen2021lyapunov} derived a convergence rate after some quantity similar to \(K_{\mix}\) iterations, and~\Cite{li2020sample} derived a sample complexity after some time related to the mixing time. For \(k \leq K_{\mix}\) , we can ensure the iterate remains bounded, and after \( k \geq K_{\mix}\), we have \(\mathcal{O}\left(\sqrt{\frac{\tau^{\mix}(\alpha_k)}{k}}\right) \) convergence rate. Lastly, we state a key lemma to prove a bound on~\eqref{neq:gradient_bias} with \(\mathcal{O}\left(\frac{\tau^{\mix}(\alpha_k)}{k}\right) \).


\begin{lemma}[Control of coupling~\Cite{bhandari2018finite}]\label{lem:control_coupling}
For some \( k\geq 0\), and \(\tau > 0 \), consider two random variables \(X\) and \(Y\) such that
\begin{equation*}
    X \to (s_k,a_k) \to (s_{k+\tau},a_{k+\tau}) \to Y.
\end{equation*}
Let the underlying Markov chain, $\{ (s_k ,a_k )\} _{k = 0}^\infty$, satisfies~\Cref{assmp:irreducible_aperiodic}. Moreover, let us construct independent random variables $X'$ and $Y'$ drawn from the
marginal distributions of $X$ and $Y$, respectively, i.e., $X' \sim \mathbb{P}(X =  \cdot )$ and $Y' \sim \mathbb{P}(Y =  \cdot )$ so that 
\begin{align*}
\mathbb{P}(X'=\cdot,Y'=\cdot ) = \mathbb{P}(X'=\cdot)\mathbb{P}(Y'=\cdot) = \mathbb{P}(X=\cdot)  \mathbb{P}(Y=\cdot). 
\end{align*}
Then, for any function \(f:\mathcal{X}\times \mathcal{Y}\to\mathbb{R}\) such that \(\sup_{(x,y)\in\mathcal{X} \times \mathcal{Y}} |f(x,y)| < \infty \), we have
\begin{equation*}
    | \mathbb{E} [f(X,Y)] - \mathbb{E}[f(X',Y')]| \leq 2 \sup_{(x,y)\in\mathcal{X}\times \mathcal{Y}} |f(x,y)| (m\rho^{\tau}).
\end{equation*}
\end{lemma}


\subsection{Bounding the crossing term \(\mathbb{E}[\xi_k(z_k,Q_k)]\) }\hfill\\
With our assumptions in place, we are now ready to bound the crossing term, \(\mathbb{E}[\xi_k(z_k,Q_k)]\), provided that the iterate \(z_k\) is bounded, which is required because the crossing term is a function of Markovian noise and \(z_k\).
In this subsection, we also need to establish the boundedness of \(Q^L_k - Q^* \). As in~\Cref{assm:Q_0_reward_bound}, we assume that \(||Q^L_0||_{\infty} \leq 1 \). Note that it is always possible because \(||Q_0||_{\infty} \leq 1 \), and we can always set $Q_0 = Q^L_0$ without loss of generality. 

\begin{lemma}[Boundedness of \(Q^L_k-Q^*\) and \(z_k\)]\label{lem:bound_z_k}
Suppose that \(||Q^L_0||_{\infty} \leq 1 \) holds. Then, the following statements hold true:
 \begin{enumerate}[1)]
     \item  For all \(k\geq 0 \), we have
\begin{align*}
    ||Q_k^L-Q^*||_{\infty} \leq \frac{4}{(1-\gamma)^2 d_{\min}} .
\end{align*}
     \item From item 1), for all \(k\geq 0 \) we have,   
     \begin{align}
    ||z_k||_{\infty} \leq \frac{8(|\mathcal{S}||\mathcal{A}|)^{\frac{1}{2}}}{(1-\gamma)^{\frac{5}{2}}d_{\min}^{\frac{3}{2}}} \nonumber.
\end{align}
 \end{enumerate}
\end{lemma}

\begin{proof}
We first prove the boundedness of \(Q_k^L-Q^*\) by an induction argument. 
For \(k=0\), we have
 \begin{equation*}
     ||Q^L_0 -Q^*||_{\infty} \leq \frac{4}{(1-\gamma)^2 d_{\min}} ,
 \end{equation*}
because
 \begin{align*}
     ||Q^L_0 - Q^*||_{\infty} &\leq ||Q^L_0 ||_{\infty} + ||Q^*||_{\infty}\\
     &\leq 1 + \frac{1}{1-\gamma} \\
     &= \frac{2-\gamma}{1-\gamma} \leq \frac{4}{(1-\gamma)^2 d_{\min}},
 \end{align*}   
 where the last inequality follows from~\Cref{lem:q_k_bound}. Next, suppose that \(||Q^L_k - Q^*||_{\infty} \leq \frac{4}{(1-\gamma)^2 d_{\min}} \) holds for some \(k\geq 0\). Then, we have
\begin{align}
    ||Q^L_{k+1}-Q^*||_{\infty} &\leq ||A_{Q^*,k}||_{\infty} ||Q^L_k -Q^*||_{\infty} +\alpha_k ||w_k||_{\infty} \nonumber \\
                           &\leq (1-d_{\min}(1-\gamma) \alpha_k) ||Q^L_k -Q^*||_{\infty} +\alpha_k \frac{4}{1-\gamma}  \nonumber \\
                           &\leq \frac{4}{(1-\gamma)^2d_{\min}} \nonumber,
\end{align} 
where the first inequality follows from the update of lower comparison system~\eqref{eq:lower_comparison_update}, and the second inequality comes from the bound on \(w_k\) in~\Cref{lem:markovian_noise_bound} and \(A_{Q^*,k}\) in~\Cref{lem:spectral_radius_of_A}.
Then, the proof is completed by induction. Next, the boundedness of \(z_k\) naturally follows from the coordinate transform in~\eqref{eq:coordinate_transform_sa}. In particular, it follows from the inequalities
\begin{align}
        ||z_k||_{\infty} &\leq ||G^{1/2}||_{\infty} ||Q^L_k-Q^*||_{\infty}\leq  \frac{8(|\mathcal{S}||\mathcal{A}|)^{\frac{1}{2}}}{(1-\gamma)^{\frac{5}{2}}d_{\min}^{\frac{3}{2}}} \nonumber,
\end{align}
where the second inequality follows from~\Cref{lem:G_upper_bound} to bound $||G^{1/2}||_{\infty}$. This completes the proof.
\end{proof} 
With the boundedness of \(z_k\), we can bound the crossing term as follows.
\begin{lemma}[Bound on \(\xi_k(z_k,Q_k)\) and Lipschitzness]\label{lem:bdd_xi}\label{lem:lipschitz-xi}\label{lem:lipschitz-xi}

\begin{enumerate}[1)]
    \item The crossing term \(\xi_k(z_k,Q_k)\) at \(k\)-th iterate is always bounded as follows:
   \begin{align}
       |\xi_k(z_k,Q_k)| \leq \frac{32 (||\mathcal{S}||\mathcal{A}|)^3}{(1-\gamma)^4d_{\min}^2 } \nonumber.
   \end{align}
   \item  For any \(k \geq 0\), \(\xi_k(z,Q) \) is Lipschitz with respect to \( (z,Q) \), i.e.,
\begin{align*}
     |\xi_k (z,Q) - \xi_k (z',Q')| \leq \frac{16(|\mathcal{S}||\mathcal{A}|)^3}{(1-\gamma)^3d_{\min}^2} ||Q-Q'||_{\infty} + \frac{4(|\mathcal{S}||\mathcal{A}|)^{\frac{3}{2}}}{(1-\gamma)^{\frac{3}{2}}d_{\min}^{\frac{1}{2}}}||z-z'||_{\infty} \nonumber
\end{align*}
\end{enumerate}
\end{lemma}

The proof is given in Appendix~\Cref{app:lem:bdd_xi}. Using above results, we are now in position to provide a bound on the crossing term, \(\mathbb{E}[\xi_k(z_{k},Q_{k})]\). Before proceeding further, we will first obtain a bound on the delayed term, \(\mathbb{E}[\xi_k(z_{k-\tau},Q_{k-\tau})]\), and then derive a bound of the crossing term, \(\mathbb{E}[\xi_k(z_{k},Q_{k})]\). 
\begin{lemma}[Boundedness of {$\mathbb{E}[\xi_k(z_{k-\tau},Q_{k-\tau})]$} ]\label{lem:boundedness_of_ex_xi_k_tau}
For any \(\tau\) and \(k\) such that \( 0 \leq \tau \leq k\), there exist \(m>0\) anr \(0<\rho<1\) such that
\begin{align*}
\mathbb{E}[\xi_k(z_{k-\tau},Q_{k-\tau})]  &\leq  \frac{96(|\mathcal{S}||\mathcal{A}|)^{3}}{(1-\gamma)^4 d_{\min}^2} (m\rho^{\tau}) \nonumber .
\end{align*}
\end{lemma}

\begin{proof}
Using some simple decomposition, we can get
\begin{align}
& |{\mathbb E}[\xi _k (z_{k - \tau } ,Q_{k-\tau} )]| \nonumber\\
&= |{\mathbb E}[\xi _k (z_{k - \tau } ,Q_{k-\tau} )]- {\mathbb E}[\xi (z_{k - \tau } ',Q_{k-\tau}',o_k')] + {\mathbb E}[\xi (z_{k - \tau } ',Q_{k-\tau}',o_k')]| \nonumber\\
&\leq |{\mathbb E}[\xi (z_{k - \tau } ,Q_{k-\tau},o_k )] - {\mathbb E}[\xi (z_{k - \tau } ',Q_{k-\tau}',o_k')]| +|{\mathbb E}[\xi (z_{k - \tau } ',Q_{k-\tau}',o_k')]|, \label{ineq:decomposition_coupling}
\end{align}
where the first inequality is due to triangle inequality. The first term in the above inequality can be bounded with the coupling lemma in~\Cref{lem:control_coupling}, and the second term can be bounded with the help of independent coupling construction and ergodic assumption in~\Cref{assmp:irreducible_aperiodic}.

First, we will bound \(|{\mathbb E}[\xi (z_{k - \tau } ,Q_{k-\tau},o_k )] - {\mathbb E}[\xi (z_{k - \tau } ',Q_{k-\tau}',o_k')]|\) in~\eqref{ineq:decomposition_coupling}. To apply~\Cref{lem:control_coupling}, first note the relations
\begin{align}
 (z_{k-\tau},Q_{k-\tau}) \to (s_{k-\tau},a_{k-\tau}) \to (s_k,a_k) \to o_k, \nonumber
\end{align}
where the arrow notation is defined in~\Cref{def:conditional-independence}, and $o_k:= (s_k,a_k, r_{k+1}, s_{k+1})$. Now, let us construct the coupling distribution of \((z_{k-\tau}',Q_{k-\tau}')\) and \(o'_k\) drawn independently from some marginal distributions of \((z_{k-\tau},Q_{k-\tau})\) and \(o_k\), respectively, i.e., $(z_{k - \tau}^{\prime}   ,Q_{k - \tau  }^\prime ) \sim {\mathbb P}((z_{k - \tau } ,Q_{k - \tau } ) =  \cdot )$ and $o_k ' \sim {\mathbb P}(o_k  =  \cdot )$. From the fact that \(\xi_k\) is bounded, which is given in the first statement of~\Cref{lem:bdd_xi}, we can now apply~\Cref{lem:control_coupling}, leading to \begin{align}
|{\mathbb E}[\xi (z_{k-\tau} ,Q_{k-\tau},o_k )] - {\mathbb E}[\xi  (z_{k-\tau}',Q'_{k-\tau},o_k ')]| \leq  \frac{64(|\mathcal{S}||\mathcal{A}|)^{3}}{(1-\gamma)^4 d_{\min}^2}  m\rho ^\tau  \label{eq:1}.
\end{align}
Next, we will bound \( |\mathbb{E}[\xi(z'_{k-\tau},Q'_{k-\tau},o'_k)]|\) in~\eqref{ineq:decomposition_coupling} from the construction of coupling that \((z'_{k-\tau},Q'_{k-\tau})\) and \(o'_k\) are independent, and ergodic assumption from~\Cref{assmp:irreducible_aperiodic}.
Using the tower property of expectation, i.e., \(\mathbb{E}[X]= \mathbb{E}[\mathbb{E}[X|Y]]\), convexity of \(|\cdot|\), and Jensen's inequality, we have 
\begin{align}
    |\mathbb{E}[\xi(z'_{k-\tau},Q'_{k-\tau},o'_k))]| &= |\mathbb{E}[\mathbb{E}[\xi(z'_{k-\tau},Q'_{k-\tau},o'_k)|z_{k-\tau}',Q_{k-\tau}']]| \nonumber\\
                &\leq \mathbb{E}[|\mathbb{E}[\xi(z'_{k-\tau},Q'_{k-\tau},o'_k)|z_{k-\tau}',Q_{k-\tau}']|]. \label{ineq:tower_1}
\end{align}

Now, our aim is to bound \(|\mathbb{E}[\xi(z'_{k-\tau},Q'_{k-\tau},o'_k)|z_{k-\tau}',Q_{k-\tau}']|\) in the last inequality. Denoting the noise term in~\eqref{def:w_k_markov} with the observation \(o'_k\) as \(w_{k}'(Q_{k-\tau}')\), we first bound the noise term. Let us define \(D_{\mu_k}\) to denote the state-action distribution at \(k\)-th time step as follows:
\begin{align*}
(e_s  \times e_a )^{\top} D_{\mu_k}  (e_s  \times e_a ) = \mu _k  (s,a) = {\mathbb P}[s_k  = s,a_k  = a|\beta ],
\end{align*}
where \(\mu_k\) is defined in~\eqref{eq:mu_k}. The noise term \( w_{k}'(Q_{k-\tau}')\) defined in~\eqref{def:w_k_markov} can be bounded as follows:
\begin{align}
    ||\mathbb{E}[w_{k}'(Q_{k-\tau}')|z_{k-\tau}',Q_{k-\tau}']||_{\infty}&\leq  ||D_{\mu_k}-D_{\infty}||_{\infty} ||R+\gamma P\Pi_{Q_{k-\tau}}Q_{k-\tau}-Q_{k-\tau}||_{\infty}  \nonumber \\
 &\leq  \frac{2}{1-\gamma} ||D_{\mu_k}-D_{\infty}||_{\infty} \nonumber\\
 &=  \frac{2}{1-\gamma} \max_{(s,a)\in \mathcal{S}\times \mathcal{A
 }} |\mu_k(s,a)-\mu_{\infty}(s,a) | \nonumber \\
  &=  \frac{2}{1-\gamma} d_{\tv}(\mu_k,\mu_{\infty})  \nonumber \\
 &\leq \frac{2}{1-\gamma}m \rho^{k},  \label{ienq:w_k_exp}
\end{align}
where the first  inequality follows from the fact that \(o'_k\) is independent with \(Q'_{k-\tau}\), and the second inequality follows from~\Cref{assm:Q_0_reward_bound} and~\Cref{lem:q_k_bound}. In particular, the first equality comes from the definition of the infinity norm, and the second equality follows from definition of the total variation distance in~\Cref{def:tv}. The last inequality follows from the ergodicity assumption of the Markov chain in~\Cref{assumption:mixing}.

With the above result, we can now bound \( |\mathbb{E}[\xi(z'_{k-\tau},Q'_{k-\tau},o'_k)|z_{k-\tau}',Q_{k-\tau}']|\) as follows:
\begin{align*}
    |\mathbb{E}[\xi(z'_{k-\tau},Q'_{k-\tau},o'_k) &|z_{k-\tau}',Q_{k-\tau}']| \\
    &= |z_{k-\tau}^{\top\prime}X_k^{\top} G^{1/2} \mathbb{E}[  w'_k(Q'_{k-\tau})|z_{k-\tau}',Q_{k-\tau}']|\\
    &\leq |\mathcal{S}| |\mathcal{A}|||X_k z_{k-\tau}'||_2 ||G^{1/2}||_2 ||\mathbb{E}[  w'_k(Q'_{k-\tau})|z_{k-\tau}',Q_{k-\tau}']||_{\infty}\\
    &\leq  \frac{16(|\mathcal{S}||\mathcal{A}|)^3}{(1-\gamma)^4d_{\min}^2} ||\mathbb{E}[  w_k'(Q_{k-\tau}')|z_{k-\tau}',Q_{k-\tau}'] ||_{\infty}\\
    &\leq \frac{32(|\mathcal{S}||\mathcal{A}|)^3}{(1-\gamma)^4d_{\min}^2} m \rho^k,
\end{align*}
where the first equality is due to the definition of the crossing term in~\eqref{neq:gradient_bias}. The first inequality is due to H\"older's inequality, and the second inequality follows from~\Cref{lem:bound_z_k} and~\Cref{lem:G_upper_bound}. Applying~\eqref{ienq:w_k_exp}, we obtain the last inequality. Combining the above result with~\eqref{ineq:tower_1} leads to
\begin{align}
    |\mathbb{E}[\xi(z'_{k-\tau},Q'_{k-\tau},o'_k))]| \leq \frac{32(|\mathcal{S}||\mathcal{A}|)^3}{(1-\gamma)^4d_{\min}^2} m \rho^k .\label{ineq:abs_xi_k_couping}
\end{align}

Now, we are ready to finish the proof. Bounding each term in~\eqref{ineq:decomposition_coupling} with~\eqref{eq:1} and~\eqref{ineq:abs_xi_k_couping}, we have
\begin{align*}
 |{\mathbb E}[\xi _k (z_{k - \tau } ,Q_{k-\tau} )]| \le \frac{96(|\mathcal{S}||\mathcal{A}|)^{3}}{(1-\gamma)^4 d_{\min}^2} m\rho ^\tau.
\end{align*}
This completes the proof. 
\end{proof}

The above lemma states that if \(\tau\), the time difference between the new observation \(o_k\) and previous iterate \(Q_{k-\tau}\), is large, then the Markovian noise term \(w(o_k,Q_{k-\tau})\) inherits the geometric mixing property of the Markov chain. Now, combining the above result with Lipschitzness of \(\xi_k (z_k,Q_k)\), we can bound the crossing term, \(\mathbb{E}[\xi_k(z_k,Q_k)] \), as follows.

\begin{lemma}[Boundedness of {$\mathbb{E}[\xi_k (z_k,Q_k)]$} ]\label{lem:boundedness_of_ex_xi}

For \( k \geq K_{\mix}\), expectation of the cross term can be bounded as follows:
\begin{align}
    \mathbb{E}[\xi_k (z_k,Q_k)] &\leq \left( \frac{96(|\mathcal{S}||\mathcal{A}|)^{3}}{(1-\gamma)^4 d_{\min}^2} +  \frac{64(|\mathcal{S}||\mathcal{A}|)^{\frac{13}{4}}}{(1-\gamma)^4 d_{\min}^2} \tau^{\mix}(\alpha_k)\right)\alpha_{k-\tau^{\mix}(\alpha_k)} . \nonumber
\end{align}

\end{lemma}
\begin{proof}
From Lipschitzness of \(\xi_k(z_k,Q_k)\) in~\Cref{lem:lipschitz-xi}, we have
\begin{align}
        \xi_k (z_k,Q_k) -  \xi_k (z_{k-\tau},Q_{k-\tau}) 
        &\leq | \xi_k (z_k,Q_k) -  \xi_k (z_{k-\tau},Q_{k-\tau})|\nonumber\\
        &\leq \frac{16(|\mathcal{S}||\mathcal{A}|)^3}{(1-\gamma)^3d_{\min}^2}||Q_k-Q_{k-\tau}||_{\infty} \nonumber\\
        &+  \frac{2(|\mathcal{S}||\mathcal{A}|)^{\frac{3}{2}}}{(1-\gamma)^{\frac{5}{2}}d_{\min}^{\frac{1}{2}}}||z_k-z_{k-\tau}||_{\infty} \label{ineq:xi_1}.
\end{align}
In the sequel, \(||Q_k-Q_{k-\tau}||_{\infty}\) and \(||z_k-z_{k-\tau}||_{\infty}\) will be bounded in terms of step-size. From the update of \(z_i\) in~\eqref{eq:coordinate_transform_sa}, for \(i\geq 0\), we have
\begin{align}
    ||z_{i+1} - z_i ||_{\infty} &= || \alpha_i B z_i + \alpha_i G^{1/2}w_i||_{\infty} \nonumber \\
    &\leq (||B||_{\infty} ||z_i||_{\infty} + ||G^{1/2}||_{\infty }||w_i||_{\infty})\alpha_i \nonumber \\
    &\leq  \left( ||B||_{2}||z_i||_{\infty} + ||G^{1/2}||_{2}  \frac{4}{1-\gamma} \right)\alpha_i  \nonumber \\
    &\leq \left((1-\gamma)d_{\min}(|\mathcal{S}||\mathcal{A}|)^{\frac{5}{4}} \frac{8(|\mathcal{S}||\mathcal{A}|)^{\frac{1}{2}}}{(1-\gamma)^{\frac{5}{2}}d_{\min}^{\frac{3}{2}}} + \sqrt{\frac{|\mathcal{S}||\mathcal{A}|}{2(1-\gamma)d_{\min}}} \frac{4}{1-\gamma}\right)\alpha_i \nonumber\\
    &\leq  \frac{16(|\mathcal{S}||\mathcal{A}|)^{\frac{7}{4}}}{(1-\gamma)^{\frac{3}{2}}d_{\min}^{\frac{1}{2}}}\alpha_i  \label{ineq:z-z},
\end{align}
where the second inequality follows from bounding \(w_i\) in~\Cref{lem:markovian_noise_bound}, and the third inequality is obtained by bounding each terms with~\Cref{lem:B_upper_bound} and~\Cref{lem:bound_z_k}. 
Similarly for \(Q_i\), we have
\begin{align}
    ||Q_{i+1}-Q_i||_{\infty} &= \alpha_i||(e_{a_k} \otimes e_{s_k} ) (e_{a_k} \otimes e_{s_k})^{\top} R + \gamma (e_{a_k} \otimes e_{s_k})(e_{s_{k+1}})^{\top}\Pi_{Q_k}Q_k \nonumber\\
    &- (e_{a_k} \otimes e_{s_k})(e_{a_k}\otimes e_{s_k})^{\top} Q_k||_{\infty}  \nonumber\\
    &\leq \frac{2-\gamma}{1-\gamma} \alpha_i . \label{ineq:Q-Q^*}
\end{align}
For any \(k\geq 1\) and \(\tau \) such that  \( 1 \leq \tau \leq k\), summing up~\eqref{ineq:z-z} and~\eqref{ineq:Q-Q^*} from \(i=k -\tau\) to \(k-1\), respectively, we get
\begin{align}
    ||z_k - z_{k-\tau}||_{\infty} &\leq \sum^{k-1}_{i=k-\tau} || z_{i+1}- z_i||_{\infty} \leq \frac{16(|\mathcal{S}||\mathcal{A}|)^{\frac{7}{4}}}{(1-\gamma)^{\frac{3}{2}}d_{\min}^{\frac{1}{2}}}  \sum^{k-1}_{i=k-\tau} \alpha_i  \leq \frac{16(|\mathcal{S}||\mathcal{A}|)^{\frac{7}{4}}}{(1-\gamma)^{\frac{3}{2}}d_{\min}^{\frac{1}{2}}} \tau \alpha_{k-\tau} \nonumber.
\end{align}
The first inequality follows from triangle inequality. The second inequality follows from~\eqref{ineq:z-z}. The last inequality follows from the fact that \(\alpha_k\) is decreasing sequence. Similarly, using~\eqref{ineq:Q-Q^*}, we have
\begin{align}
    ||Q_k - Q_{k-\tau}||_{\infty} &\leq \sum^{k-1}_{i=k-\tau} || Q_{i+1}- Q_i||_{\infty} \leq \frac{2-\gamma}{1-\gamma}  \sum^{k-1}_{i=k-\tau} \alpha_i \leq \frac{2-\gamma}{1-\gamma}\tau \alpha_{k-\tau}. \nonumber
\end{align}
Therefore, combining the above results with~\eqref{ineq:xi_1} leads to
    \begin{align}\label{ineq:xi_2}
        \xi_k (z_k,Q_k) -  \xi_k (z_{k-\tau},Q_{k-\tau}) 
        &\leq \frac{32(|\mathcal{S}||\mathcal{A}|)^{3}}{(1-\gamma)^4 d_{\min}^2} \tau\alpha_{k-\tau}
        + \frac{4(|\mathcal{S}||\mathcal{A}|)^{\frac{13}{4}}}{(1-\gamma)^4d_{\min}}\tau\alpha_{k-\tau} \nonumber\\
        &\leq  \frac{64(|\mathcal{S}||\mathcal{A}|)^{\frac{13}{4}}}{(1-\gamma)^4 d_{\min}^2} \tau \alpha_{k-\tau}.
    \end{align}
Next, we will bound \( \mathbb{E}[\xi_k (z_k,Q_k)]\). In particular, taking total expectations on~\eqref{ineq:xi_2} yields
\begin{align}
       \mathbb{E}[ \xi_k (z_k,Q_k)] \leq \mathbb{E}[\xi_k (z_{k-\tau},Q_{k-\tau})] +   \frac{64(|\mathcal{S}||\mathcal{A}|)^{\frac{13}{4}}}{(1-\gamma)^4 d_{\min}^2}  \tau \alpha_{k-\tau} \nonumber.
\end{align}

Since the above inequality is satisfied for any \( 1\leq \tau \leq k\), for \( k \geq K_{\mix}\), we can set \(\tau=\tau^{\mix}(\alpha_k)\) from the definition of $\tau^{\mix}(\alpha_k)$. Now, we have
\begin{align}
    \mathbb{E}[\xi_k (z_k,Q_k)] 
    &\leq \frac{96(|\mathcal{S}||\mathcal{A}|)^{3}}{(1-\gamma)^4 d_{\min}^2} (m\rho^{\tau^{\mix}(\alpha_k)}) +
     \frac{64(|\mathcal{S}||\mathcal{A}|)^{\frac{13}{4}}}{(1-\gamma)^4 d_{\min}^2}  \tau^{\mix}(\alpha_k) \alpha_{k-\tau^{\mix}(\alpha_k)} \nonumber \\
    &\leq  \frac{96(|\mathcal{S}||\mathcal{A}|)^{3}}{(1-\gamma)^4 d_{\min}^2} \alpha_k +  \frac{64(|\mathcal{S}||\mathcal{A}|)^{\frac{13}{4}}}{(1-\gamma)^4 d_{\min}^2}  \tau^{\mix}(\alpha_k) \alpha_{k-\tau^{\mix}(\alpha_k)}  \nonumber\\
    &\leq \left( \frac{96(|\mathcal{S}||\mathcal{A}|)^{3}}{(1-\gamma)^4 d_{\min}^2} +  \frac{64(|\mathcal{S}||\mathcal{A}|)^{\frac{13}{4}}}{(1-\gamma)^4 d_{\min}^2} \tau^{\mix}(\alpha_k)\right)\alpha_{k-\tau^{\mix}(\alpha_k)} \nonumber,
\end{align}
where the first inequality follows from~\Cref{lem:boundedness_of_ex_xi_k_tau}, the second inequality comes from the definition of mixing time in~\eqref{def:mixing_time}, and the last inequality follows from the fact that \(\alpha_k\) is a decreasing sequence.
\end{proof}

\subsection{Finite time analysis under Markovian noise}\hfill\\
Now, we turn to the statement of our main results. The proof logic follows similar arguments as in the i.i.d. setting with few modifications. First, we begin the overall proof by deriving the convergence rate of \(z_k\), which corresponds to the lower comparison system with a coordinate transformation. Then, as a next step, we will consider the difference system to obtain a convergence rate of the original system. In the sequel, a generic constant \(C > 0\) will be used for unimportant values that change line by line.

\begin{proposition}[Convergence rate of \(z_k\)]\label{prop:convergent_rate_z_k_markovian}
Consider the step-size in~\eqref{eq:step_size} and Markovian observation model. For any \(k\geq K_{\mix}\), we have
    \begin{align}
    \mathbb{E}[V(z_k)] \leq  \frac{C_2}{k+\xi}+\frac{C (|\mathcal{S}||\mathcal{A}|)^{\frac{21}{4}}}{(1-\gamma)^6 d_{\min}^4} \frac{\tau^{\mix}(\alpha_k)}{k+\xi}  \nonumber,
    \end{align}
     where \(V(x)= x^{\top}x\) and \( C_2 := C \max \left\{  \frac{(|\mathcal{S}||\mathcal{A}|)^{\frac{11}{2}}d_{\max}^2}{(1-\gamma)^3d_{\min}^3}||Q^L_0 -Q^*||^2_2 ,
    \frac{(|\mathcal{S}| |\mathcal{A}|)^{5}}{(1-\gamma)^6d_{\min}^4} 
    \right\}\).
\end{proposition}
\begin{proof}
We will follow similar lines as in the proof of~\Cref{prop:lower_z_convergence_rate}. For simplicity of the induction proof, let \( C_1 := \max \left\{ \frac{8(|\mathcal{S}||\mathcal{A}|)^{\frac{11}{2}}d_{\max}^2}{(1-\gamma)^3d_{\min}^3}||Q^L_0 -Q^*||^2_2 ,
   \frac{72(|\mathcal{S}||\mathcal{A}|)^4}{(1-\gamma)^5d_{\min}^3}
    \right\} \). Plugging the term \(z_{k+1}=X_kz_k+\alpha_k G^{1/2}w_k\) into \(V(z_{k+1})\), we have the following relation: 
\begin{align*}
    &\mathbb{E}[V(z_{k+1})|\mathcal{F}^L_k] \\
    &= \mathbb{E}[(z_k)^{\top}X_k^{\top}X_k(z_k)+ 2\alpha_k z_k^{\top}X_k^{\top}G^{1/2}w_k + \alpha_k^2 w_k^{\top}Gw_k| \mathcal{F}^L_k] \nonumber\\
    &\leq (1-\beta\alpha_k)\mathbb{E}[V(z_k)] + 2\alpha_k \mathbb{E}[\xi_k(z_k,Q_k)\mid\mathcal{F}^L_k] + \lambda_{\max}(G) \frac{\theta^2}{(k+\xi)^2} \mathbb{E}[||w_k||^2_2 \mid \mathcal{F}_k^L] \\
    &\leq \left(1- \frac{2}{k+\xi} \right) \mathbb{E}[V (z_k)]  +\frac{C_1}{(k+\xi)^2} + 2\alpha_k \mathbb{E}[\xi_k (z_k,Q_k)\mid \mathcal{F}_k^L] ,
\end{align*}
where \(\mathcal{F}^L_k\) is a \(\sigma\)-field induced by \(\{ (s_i,a_i,s_{i+1},r_i, Q_i,Q_i^L)\}_{i=0}^k\), the first inequality follows from bound on \(||X_k||_2\) in~\Cref{lem:bound on singular value of X_k}, and positive definiteness of \(G\), the second inequality follows from the step-size condition in~\Cref{lem:bound on singular value of X_k}, and collecting the bound on \(\lambda_{\max}(G) \) in~\Cref{lem:G_upper_bound}, \(w_k\) in~\Cref{lem:markovian_noise_bound}, and \(\theta\) in~\Cref{lem:theta_bound}. Taking total expectations, it follows that
\begin{align}
        \mathbb{E}[V(z_{k+1})] &\leq \left(1- \frac{2}{k+\xi} \right) \mathbb{E}[V (z_k)] + \frac{C_1}{(k+\xi)^2} + 2\alpha_k \mathbb{E}[\xi_k (z_k,Q_k)] \nonumber,
\end{align}
The main difference compared to~\eqref{eq:v_low_expand} in~\Cref{prop:lower_z_convergence_rate} is that we have the non-zero term \( \mathbb{E}[\xi_k(z_k,Q_k)] \). Now, bounding \(\mathbb{E}[\xi_k(z_k,Q_k)]\) leads to
\begin{align}
          \mathbb{E}[V(z_{k+1})] &\leq \left(1- \frac{2}{k+\xi} \right) \mathbb{E}[V (z_k)] + \frac{C_1}{(k+\xi)^2} \nonumber\\
    &+ 2\frac{\theta}{k+\xi} \left( \frac{96(|\mathcal{S}||\mathcal{A}|)^{3}}{(1-\gamma)^4 d_{\min}^2} +  \frac{64(|\mathcal{S}||\mathcal{A}|)^{\frac{13}{4}}}{(1-\gamma)^4 d_{\min}^2} \tau^{\mix}(\alpha_k)\right) \alpha_{k-\tau^{\mix}(\alpha_k)} \nonumber  \\
         &\leq \left(1- \frac{2}{k+\xi} \right) \mathbb{E}[V (z_k)] + \frac{C_1}{(k+\xi)^2} \nonumber\\
    &+ 2\frac{\theta}{k+\xi} \left( \frac{96(|\mathcal{S}||\mathcal{A}|)^{3}}{(1-\gamma)^4 d_{\min}^2} +  \frac{64(|\mathcal{S}||\mathcal{A}|)^{\frac{13}{4}}}{(1-\gamma)^4 d_{\min}^2} \tau^{\mix}(\alpha_k)\right)\frac{2\theta}{k+\xi} \nonumber  \\
          &=  \left(1- \frac{2}{k+\xi} \right) \mathbb{E}[V (z_k)] +  \frac{C_2}{(k+\xi)^2} + \frac{256\theta^2 (|\mathcal{S}||\mathcal{A}|)^{\frac{13}{4}}}{(1-\gamma)^4 d_{\min}^2}\frac{\tau^{\mix}(\alpha_k)}{(k+\xi)^2} \nonumber \\
          &\leq  \left(1- \frac{2}{k+\xi} \right) \mathbb{E}[V (z_k)] +  \frac{C_2}{(k+\xi)^2} + \frac{C (|\mathcal{S}||\mathcal{A}|)^{\frac{21}{4}}}{(1-\gamma)^6 d_{\min}^4} \frac{\tau^{\mix}(\alpha_k)}{(k+\xi)^2} \nonumber,
\end{align}
where the first inequality is obtained by bounding  \(\mathbb{E}[\xi_k(z_k,Q_k)]\) by~\Cref{lem:boundedness_of_ex_xi_k_tau}. The second inequality is due to the definition of \(K_{\mix}\) in~\eqref{def:K}. The third inequality follows from bound on \(\theta\) in~\Cref{lem:theta_bound} given in Appendix~\Cref{app:sec:step_size}.

Applying the same logic of induction as in the proof of~\Cref{prop:lower_z_convergence_rate}, we have
\begin{align}
    \mathbb{E}[V(z_k)] \leq \frac{C_2}{k+\xi}+\frac{C (|\mathcal{S}||\mathcal{A}|)^{5}}{(1-\gamma)^6 d_{\min}^4} \frac{\tau^{\mix}(\alpha_k)}{k+\xi} . \nonumber
\end{align}
This completes the proof.
\end{proof}
From~\eqref{def:z_k}, we can derive the final iterate convergence rate for \(  \mathbb{E}[|| Q^L_k-Q^* ||_{\infty}]\) corresponding to the lower comparison system.
\begin{theorem}\label{thm:markovian_lower_system_convergence_rate}
For any \(k \geq K_{\text{mix}}\), we have
\begin{align}
     \mathbb{E}[|| Q^L_k-Q^* ||_{\infty}] &\leq \frac{
     C^L}{\sqrt{k+\xi}} + \frac{C(|\mathcal{S}||\mathcal{A}|)^{\frac{25}{8}}}{(1-\gamma)^{3} d_{\min}^{2}} \frac{\sqrt{\tau^{\mix}(\alpha_k)}}{\sqrt{k+\xi}} \nonumber,
\end{align}
where \(C^L:=  C \max \left\{ \frac{(|\mathcal{S}||\mathcal{A}|)^3d_{\max}^{\frac{3}{2}}}{(1-\gamma)^{\frac{3}{2}}d_{\min}^{\frac{3}{2}}}||Q^L_0 -Q^*||_2,
    \frac{ (|\mathcal{S}| |\mathcal{A}|)^{\frac{11}{4}}}{(1-\gamma)^{3}d_{\min}^{2}} 
    \right\}  \).
\end{theorem}
\begin{proof}

Noting that \( Q^L_k-Q^* = G^{-1/2} z_{k}\) and from~\Cref{prop:convergent_rate_z_k_markovian}, we have

\begin{align*}
    \mathbb{E}[|| Q^L_k-Q^* ||^2_2]
    &\leq \lambda_{\max}(G^{-1}) \left( C \max \left\{  \frac{(|\mathcal{S}||\mathcal{A}|)^{\frac{11}{2}}d_{\max}^2}{(1-\gamma)^3d_{\min}^3}||Q^L_0 -Q^*||^2_2 ,
    \frac{(|\mathcal{S}| |\mathcal{A}|)^{5}}{(1-\gamma)^6d_{\min}^4} 
    \right\} \frac{1}{k+\xi} \right. \\
    &\left.+\frac{C (|\mathcal{S}||\mathcal{A}|)^{\frac{21}{4}}}{(1-\gamma)^6 d_{\min}^4} \frac{\tau^{\mix}(\alpha_k)}{k+\xi} \right) \nonumber\\
    &\leq  \frac{(C^L)^2}{k+\xi}+\frac{C (|\mathcal{S}||\mathcal{A}|)^{\frac{25}{4}}}{(1-\gamma)^6 d_{\min}^4} \frac{\tau^{\mix}(\alpha_k)}{k+\xi} , \nonumber
\end{align*}
where the second inequality follows from the upper bound on \(G^{-1}\) in ~\Cref{lem:G_upper_bound}. Using Jensen's inequality as in~\Cref{thm:lower_system_convergence_rate}, the above result leads to
\begin{align}
     \mathbb{E}[|| Q^L_k-Q^* ||_{\infty}] &\leq   \frac{
     C^L}{\sqrt{k+\xi}} + \frac{C(|\mathcal{S}||\mathcal{A}|)^{\frac{25}{8}}}{(1-\gamma)^{3} d_{\min}^{2}} \frac{\sqrt{\tau^{\mix}(\alpha_k)}}{\sqrt{k+\xi}}. \nonumber
\end{align}
This completes the proof.
\end{proof}

Next, we derive a convergence rate for \(\mathbb{E}[||Q^U_k - Q^L_k||_{\infty}]\). The proof is similar to those of the i.i.d. setting: we use a bound on \(\mathbb{E}[||Q^L_k-Q^*||_{\infty}] \), and then apply an induction logic. The randomness of the \(Q^U_k-Q^L_k\) mainly relies on \( Q^L_k-Q^*\). Therefore the influence of Markovian noise in \(Q^L_k-Q^*\) directly affects the difference between the upper and lower comparison systems. That is, the convergence rate of \(\mathbb{E}[||Q^U_k -Q^L_k ||]\) follows that of \( \mathbb{E}[||Q^L_k-Q^* ||_{\infty}]\), and sample complexity becomes worse due to slow progress of the upper comparison system as seen in the i.i.d. case.

\begin{theorem}\label{thm:markovian_upper_system_convergence_rate}
For any \(k\geq K_{\mix} \), the difference between the upper and lower comparison system is bounded as follows:
\begin{equation}
    \mathbb{E}[||Q^U_k-Q^L_k||_{\infty}] \leq \frac{C^U}{\sqrt{k+\xi}}    + \frac{C(|\mathcal{S}||\mathcal{A}|)^{\frac{31}{8}}d_{\max}}{(1-\gamma)^{\frac{7}{2}} d_{\min}^{\frac{5}{2}}} \frac{\sqrt{\tau^{\mix}(\alpha_k)}}{\sqrt{k+\xi}} \nonumber,
\end{equation}
where $C^U:=  C \max \left\{ \frac{(|\mathcal{S}||\mathcal{A}|)^4d_{\max}^{\frac{5}{2}}}{(1-\gamma)^{\frac{5}{2}}d_{\min}^{\frac{5}{2}}}||Q^L_0 -Q^*||_2,
    \frac{ (|\mathcal{S}| |\mathcal{A}|)^{\frac{15}{4}}}{(1-\gamma)^{4}d_{\min}^{3}} 
    \right\} $.
\end{theorem}
\begin{proof}
As in~\Cref{thm:upper_system_convergence_rate}, we take expectation on~\eqref{eq:up_lr_diff}, and use sub-multiplicativity of the norm. 

\begin{align}
\mathbb{E}[||Q^U_{k+1}-Q^L_{k+1}||_{\infty}] &\leq 
    \left( 1- \frac{2}{k+\xi}\right) \mathbb{E} [||Q^U_k- Q^L_k ||_{\infty}] + 2\frac{\theta}{k+\xi} \gamma d_{\max} \mathbb{E}[||Q^L_k-Q^*||_{\infty}] \nonumber \\
    &\leq  \left( 1- \frac{2}{k+\xi}\right) \mathbb{E} [||Q^U_k- Q^L_k ||_{\infty}] \nonumber\\
    &+ 2\frac{\theta}{k+\xi} \gamma d_{\max} \left(
     \frac{C^L}{\sqrt{k+\xi}} + \frac{C(|\mathcal{S}||\mathcal{A}|)^{\frac{25}{8}}}{(1-\gamma)^{3} d_{\min}^{2}} \frac{\sqrt{\tau^{\mix}(\alpha_k)}}{\sqrt{k+\xi}}
    \right)\nonumber\\
    &\leq \left( 1- \frac{2}{k+\xi}\right) \mathbb{E} [||Q^U_k- Q^L_k ||_{\infty}] \nonumber\\
    &+ \frac{1}{k+\xi}  \left(
     \frac{C^U}{\sqrt{k+\xi}} +\gamma d_{\max} \frac{C(|\mathcal{S}||\mathcal{A}|)^{\frac{33}{8}}d_{\max}}{(1-\gamma)^{4} d_{\min}^{3}} \frac{\sqrt{\tau^{\mix}(\alpha_k)}}{\sqrt{k+\xi}}
    \right) \nonumber,
\end{align}
where \( C^L:=  C \max \left\{ \frac{(|\mathcal{S}||\mathcal{A}|)^3d_{\max}^{\frac{3}{2}}}{(1-\gamma)^{\frac{3}{2}}d_{\min}^{\frac{3}{2}}}||Q^L_0 -Q^*||_2,
    \frac{ (|\mathcal{S}| |\mathcal{A}|)^{\frac{11}{4}}}{(1-\gamma)^{3}d_{\min}^{2}} 
    \right\}  \).
The first inequality follows from the same logic in~\Cref{thm:upper_system_convergence_rate}. The second inequality follows from the upper bound on the lower comparison system,~\Cref{thm:markovian_lower_system_convergence_rate}. The last inequality follows from upper bound on \(\theta\) in~\Cref{lem:theta_bound} given in Appendix~\Cref{app:sec:step_size}. Applying the same induction logic in~\Cref{thm:upper_system_convergence_rate}, we have
\begin{align}
    \mathbb{E}[||Q^U_k-Q^L_k||_{\infty}] &\leq \frac{C^U}{\sqrt{k+\xi}}    + \frac{C(|\mathcal{S}||\mathcal{A}|)^{\frac{31}{8}}d_{\max}}{(1-\gamma)^{\frac{7}{2}} d_{\min}^{\frac{5}{2}}} \frac{\sqrt{\tau^{\mix}(\alpha_k)}}{\sqrt{k+\xi}} \nonumber.
\end{align}

\end{proof}
Finally, we derive the convergence rate for the original system. which follows from simple triangle inequality.
\begin{theorem}\label{thm:markovian_convergence_rate}
For \(k \geq K_{\mix}\), we have
\begin{align}
    \mathbb{E}[||Q_k-Q^*||_{\infty}] &\leq   \frac{C^*}{\sqrt{k+\xi}}   + \frac{C(|\mathcal{S}||\mathcal{A}|)^{\frac{33}{8}}d_{\max}}{(1-\gamma)^{4} d_{\min}^{3}} \frac{\sqrt{\tau^{\mix}(\alpha_k)}}{\sqrt{k+\xi}} \nonumber,
\end{align}
where \( C^*:=  C \max \left\{ \frac{(|\mathcal{S}||\mathcal{A}|)^4d_{\max}^{\frac{5}{2}}}{(1-\gamma)^{\frac{5}{2}}d_{\min}^{\frac{5}{2}}}||Q^L_0 -Q^*||_2,
    \frac{ (|\mathcal{S}| |\mathcal{A}|)^{\frac{15}{4}}}{(1-\gamma)^{4}d_{\min}^{3}} 
    \right\} \).
\end{theorem}
\begin{proof}

Using triangle inequality, we can derive the upper bound when \( k \geq K_{\mix}\),
\begin{align}
   \mathbb{E}[||Q_k-Q^*||_{\infty}] &\leq \mathbb{E}[||Q^L_k-Q^*||_{\infty}] +\mathbb{E}[||Q_k^U-Q^L_k||_{\infty}] \nonumber\\
    &\leq \frac{C^*}{\sqrt{k+\xi}}   + \frac{C(|\mathcal{S}||\mathcal{A}|)^{\frac{33}{8}}d_{\max}}{(1-\gamma)^{4} d_{\min}^{3}} \frac{\sqrt{\tau^{\mix}(\alpha_k)}}{\sqrt{k+\xi}} \nonumber,
   \end{align}
where the first inequality is due to triangle inequality, and the second inequality follows from combining the bound on the lower comparison system,~\Cref{thm:markovian_lower_system_convergence_rate} and the difference of the upper and lower comparison systems,~\Cref{thm:markovian_upper_system_convergence_rate}. 
\end{proof}
\begin{remark}
    Since we can bound the mixing time with logarithmic term by~\Cref{lem:mixing_time_bound} given in Appendix~\Cref{app:sec:sample_complexity}, we can achieve \( \mathcal{O} \left( \sqrt{\frac{\log k}{k}} \right)\) convergence rate under Markovian observation model. However, from~\Cref{lem:markovian_sample_complexity_tail_bound} given in Appendix~\Cref{app:sec:sample_complexity}, the above result leads to sub-optimal sample complexity. 
    This implies that without using mixing time in the step-size,  even though we can still get \( \mathcal{O} \left( \sqrt{\frac{\log k}{k}} \right) \) rate, the sample complexity can get extremely worse compared to the case when we use mixing time in the step-size~\Cite{li2020sample,chen2021lyapunov,qu2020finite}. 
\end{remark}

\subsection{Comparitive analysis}\label{sec:compare}
\import{./markovian}{compartive_analysis.tex}

%% file: markovian/compartive_analysis.tex
The recent paper~\Cite{li2020sample} derives sample complexity 
in terms of the probability tail bounds using the adaptive step-size
\begin{equation*} 
\alpha _k  = \min \left\{ {1,c\exp \left( {\left\lfloor {\log \left( {\frac{\log k}{{\hat d_{\min ,k} (1 - \gamma )\gamma ^2 k}}} \right)} \right\rfloor } \right)} \right\},
\end{equation*}
where $\hat{d}_{\min,k}$ is a number that is updated at every iteration, and \(c\) is some universal constant. Note that the step-size does not depend on \(d_{\min}\) or the mixing time, while it requires an additional step to update \(\hat{d}_{\min,k} \), which brings in additional complexity compared to our diminishing step-size. Moreover, their update requires a state-action dependent process, the number of times that the current state-action visits a certain state-action pair, which requires ${\cal O}(|{\cal S}||{\cal A}|)$ memory space. With the help of concentration inequality in Markov chain~\Cite{paulin2015concentration} and refined analysis based on number of visits to \((s,a)\), probability tail bounds can be obtained as well. 

Another recent paper~\Cite{chen2021lyapunov} uses a step-size that satisfies the constraint
\begin{align*}
\sum\limits^{k-t^{\mix}(\alpha_k)}_{i=k-1} \alpha_i \leq c
\end{align*}
where
\begin{align*}
    t_{\mix} (\alpha_k) = \min \left\{  k\geq 0 \middle| \sup_{(s,a)\in \mathcal{S}\times \mathcal{A}} 
    d_{TV}(\mu_k(s,a),\mu) \leq \alpha_k
    \right\},
\end{align*}
and for some constant \(c\) to derive sample complexity in terms of the expectation using analysis based on Moreau envelope~\Cite{chen2020finite}. Under ergodic assumption, it derives \(\mathcal{O}\left( \sqrt{\frac{t_{\mix}(\alpha_k)}{k}} \right)\) where \(t_{\mix}(\alpha_k)\) is upper bounded by logarithmic term.
The study in~\Cite{even2003learning} showed that extending the analysis of~\Cite{bertsekas1995neuro} and using the step-size \( \frac{1}{k}\) lead to sample complexity depending polynomially on \(\frac{1}{1-\gamma}\).
The paper~\Cite{qu2020finite} uses the diminishing step-size \(\alpha_k = \frac{h}{k+k_0} \) with 
\begin{align*}
k_0 \geq \max \left\{ 4h , \left \lceil{\log \frac{2}{d_{\min}}}\right \rceil  t^{\mix}(1/4) \right\} ,
\end{align*} 
where \(h\geq \frac{4}{d_{\min}(1-\gamma)}\) to derive sample complexity in terms of probability tail bounds. The recent progress~\Cite{qu2020finite} proved that analysis on decomposition of error term into shifted Martingale terms leads to probability tail bounds. Most of the previous and recent analysis uses step-sizes with the associated conditions that depend on the mixing time, whereas our work uses a diminishing step-size that is independent of the mixing time. Therefore, our analysis covers different scenarios. With the construction of the upper and lower comparison systems, both of which can be analyzed without injecting the mixing time into the step-size condition, we can derive the convergence rate of the original system.

%% file: conclusion/conclusion.tex
In this paper, we have derived a final iterate convergence rate of asynchronous Q-learning under i.i.d. and Markovian observation model and diminishing step-size that is independent of the mixing time. Even though our step-size does not depend on the mixing time, we can achieve \( \mathcal{O}\left(\sqrt{\frac{\log k}{k}}\right)\), which is the sharpest convergence rate under Markovian observation model. Tightening the sample complexity to the current sharpest bound, which uses a step-size independent of the mixing time, would be one of the future research directions. Furthermore, extending this framework to Q-learning variants e.g., double Q-learning~\Cite{hasselt2010double}, periodic Q-learning~\Cite{lee2020periodic}, is left as a future work.

%% file: appendix/appendix.tex


\subsection{Proof of~\Cref{lem:diag_dom_hurwtiz}}\label{app:lem:diag_dom_hurwtiz}
We first introduce Gerschgorin circle theorem which is essential in proving~\Cref{lem:T_q_hurwitz}.
\begin{lemma}[Gerschgorin circle theorem~\cite{horn2013matrix}]\label{lem:gersgorin-circle-theroem}
Let \(A  \in \mathbb{R}^{n\times n}\) be a square matrix. Gerschgorin circles are defined as 
\begin{align*}
    \left\{ z\in \mathbb{C} \middle| |z- [A]_{ii} | \leq \sum\limits_{j \ne i,j = 1}^n |[A]_{ij}| \right\} ,\quad i=1,\dots,n.
\end{align*}
The eigenvalues of \(A\) are in the union of Gerschgorin discs,
\begin{align*}
    G(A) = \bigcup^n_{i=1}  \left\{ z\in \mathbb{C} \middle| |z- [A]_{ii} | \leq \sum\limits_{j \ne i,j = 1}^n |[A]_{ij}| \right\} .
\end{align*}
\end{lemma}
Gerschgorin circle theorem states that the eigenvalues of the matrix are within circles, called Gerschgorin discs, which are defined in terms of algebraic relations and entries of the matrix $A$. Building on~\Cref{lem:gersgorin-circle-theroem}, we can prove~\Cref{lem:diag_dom_hurwtiz}, a strictly diagonally dominant matrix with negative diagonal terms is Hurwitz matirx.

To this end, first note that \(Re(\lambda_i (A)) -[A]_{ii} \leq \left.\right| \lambda_i (A)-[A]_{ii}|\), where \(\lambda_i (A)\) denotes the \(i\)-th eigenvalue of \(A\) and \( Re(\lambda_i (A))\) denotes the real part of \(\lambda_i (A)\). Now, applying~\Cref{lem:gersgorin-circle-theroem} yields
\begin{align*}
    Re(\lambda_i (A)) \leq \sum\limits_{j \ne i,j = 1}^n|[A]_{ij}| + [A]_{ii} < 0.
\end{align*}
Hence, the real part of eigenvalues of \(A\) is negative, which means that \(A\) is Hurwtiz matrix.

\subsection{Inequalities related to step-size}\label{app:sec:step_size}
Now, we introduce a useful inequality used in the proof of~\Cref{prop:lower_z_convergence_rate}. 
\begin{lemma}\label{lem:theta_bound}\label{lem:xi_bound}
For \(\theta\) and \(\xi \) defined in~\eqref{eq:theta}, we have the following  bounds:
\begin{align}
    \theta &\leq \frac{4|\mathcal{S}||\mathcal{A}|}{(1-\gamma)d_{\min}} \nonumber,\\
    \frac{8}{(|\mathcal{S}||\mathcal{A}|)^2} &\leq \xi \leq    \frac{16(|\mathcal{S}||\mathcal{A}|)^{\frac{9}{2}}d_{\max}^2}{(1-\gamma)^2d_{\min}^2} . \nonumber
\end{align}

\end{lemma}
\begin{proof}
We first prove the bound on \(\theta\) in the first statement. By the definition of \(\theta\) in~\eqref{eq:theta}, we have the bound on \(\theta\) as follows:
    \begin{align}
    \theta &\leq \frac{8}{\nu} .\nonumber
\end{align}
Next, we further bound \( \frac{8}{\nu}\) by using the definition of \(\nu\) in~\eqref{def:nu} as follows:
\begin{align}
    \frac{8}{\nu} \leq 8\max \left\{\frac{1}{(1-\gamma)d_{\min}},\frac{1}{\lambda_{\min}(G^{-1})} \right\} \nonumber.
\end{align}

Noting that the eigenvalues of \(G^{-1}\) are inverse of eigenvalues of \(G\), and \(||G||_2\) is the largest eigenvalue of $G$ (since $G$ is symmetic positive definite matrix), we have \( \lambda_{\min}(G^{-1}) = \frac{1}{||G||_2}\). Using the bound on \(G\) in~\Cref{lem:G_upper_bound} leads to
\begin{align}
    \frac{1}{\lambda_{\min}(G^{-1})}  & = ||G||_2 \leq \frac{|\mathcal{S}||\mathcal{A}|}{2(1-\gamma)d_{\min}} \nonumber.
\end{align}

Plugging this bound into the last inequality yields
\begin{align}\label{ineq:1/nu}
    \frac{1}{\nu}\leq \frac{|\mathcal{S}||\mathcal{A}|}{2(1-\gamma)d_{\min}} ,
\end{align}
which leads to the upper bound of \( \theta\).

Next, we prove the bound on \(\xi\) in the second statement. The upper bound of \(\xi\) can be derived as follows:
\begin{align}
    \xi &\leq \frac{16 ||B||^2_2}{\nu^2} \nonumber\\
       &\leq \frac{64(|\mathcal{S}||\mathcal{A}|)^{\frac{5}{2}} d_{\max}^2}{\nu^2} \nonumber\\
       &\leq \frac{16(|\mathcal{S}||\mathcal{A}|)^{\frac{9}{2}}d_{\max}^2}{(1-\gamma)^2d_{\min}^2}, \nonumber
\end{align}
where the first inequality follows from definition of \(\xi\) in~\eqref{ineq:xi}, the second inequality due to the bound on \(B\) in~\Cref{lem:B_upper_bound}, and the last inequality follows from~\eqref{ineq:1/nu}.
It remains to prove the lower bound of \(\xi\) in the second statement. It follows from the inequalities
\begin{align}
    \xi &\geq \frac{8||B||_2^2}{\nu^2}\nonumber\\
       &\geq \frac{8(1-\gamma)^2d_{\min}^2}{(|\mathcal{S}||\mathcal{A}|)^2} \frac{1}{(1-\gamma)^2d_{\min}^2} \nonumber\\
        &\geq \frac{8}{(|\mathcal{S}||\mathcal{A}|)^2},\nonumber
\end{align}
where the second inequality follows from~\Cref{lem:B_upper_bound}, and from definition of \(\nu\) which is~\eqref{def:nu}. This completes the proof. 
\end{proof}



\subsection{Proof of~\Cref{lem:q_k_bound}}\label{app:proof:q_k_bound}
Assume \( |Q_k (s_k,a_k)| \leq \frac{1}{1-\gamma}\). Then, using triangle inequality to Q-learning update in~\Cref{algo:standard-Q-learning1} and the assumption that \(||R||_{\infty}\leq1\) leads to
\begin{align}
    |Q_{k+1}(s_k,a_k)| &\leq (1-\alpha_k)\frac{1}{1-\gamma} +\alpha_k \left(1+\gamma \frac{1}{1-\gamma}\right)\\
                        &\leq \frac{1}{1-\gamma} \nonumber.
\end{align}
Therefore, using induction, we have the desired inequality.

\subsection{Proof of~\Cref{lem:w_k_bound}}\label{app:proof:w_k_bound}
To prove the first result in~\eqref{ineq:bound on w_k}, we have
\begin{align*}
    ||w_k||_{\infty} &\leq 2 + 2(1+\gamma)||Q_k||_{\infty} \leq 2+ \frac{2(1+\gamma)}{1-\gamma} = \frac{4}{1-\gamma},
\end{align*}
where the first inequality follows from applying triangle inequality to the definition of \(w_k\) in~\eqref{eq:noise_term}, and the second inequality comes from~\Cref{lem:q_k_bound}. Next, we prove the second result in~\eqref{ienq:bound_w_k_2}. For simplicity, let 
\begin{align*}
    \delta_k &:= (e_{a_k} \otimes e_{s_k} ) (e_{a_k} \otimes e_{s_k})^{\top} R + \gamma (e_{a_k} \otimes e_{s_k})(e_{s_{k+1}})^{\top}\Pi_{Q_k}Q_k \\
    &-  (e_{a_k} \otimes e_{s_k} ) (e_{a_k} \otimes e_{s_k})^{\top} ) Q_k \nonumber .
\end{align*}

Then, we have the following relation:
\begin{align}
    \mathbb{E}[w^{\top}_k w_k \mid \mathcal{F}_k] &= \mathbb{E}[\delta^{\top}_k \delta_k\mid \mathcal{F}_k] - \mathbb{E}[\delta_k^{\top} \mathbb{E}[\delta_k \mid \mathcal{F}_k] \mid \mathcal{F}_k] \\
    &- \mathbb{E}[\mathbb{E}[\delta_k \mid \mathcal{F}_k]^{\top}\delta_k \mid \mathcal{F}_k] + ||\mathbb{E}[\delta_k \mid \mathcal{F}_k]||^2_2 \nonumber\\
 &=   \mathbb{E}[||\delta_k||^2_2 \mid \mathcal{F}_k]  -  ||\mathbb{E}[\delta_k \mid \mathcal{F}_k]||^2_2 \nonumber\\
 &\leq \mathbb{E}[||\delta_k||^2_2 \mid \mathcal{F}_k].  \label{eq:w_k_var}
\end{align}

where \( \mathcal{F}_k\) is a \(\sigma\)-field induced by \(\{ (s_i,a_i,s_{i}',r_i, Q_i)\}_{i=0}^k\), the second equality follows from the fact that \( \mathbb{E}[\delta_k | \mathcal{F}_k]\) is \( \mathcal{F}_k\)-measurable, meaning that \(\mathbb{E}[\mathbb{E}[\delta_k \mid \mathcal{F}_k]\mid \mathcal{F}_k] = \mathbb{E}[\delta_k \mid \mathcal{F}_k]\). Next, we bound \( ||\delta_k||_2 \) as follows:
\begin{align*}
    ||\delta_k||_2 &\leq ||\delta_k||_1  \nonumber\\
                 &\leq  || (e_{a_k} \otimes e_{s_k} ) (e_{a_k} \otimes e_{s_k})^{\top} R||_1 + \gamma ||(e_{a_k} \otimes e_{s_k})(e_{s_k'})^{\top}\Pi_{Q_k}Q_k||_1 \nonumber\\
                 &+ \gamma ||  (e_{a_k} \otimes e_{s_k} ) (e_{a_k} \otimes e_{s_k})^{\top} ) Q_k ||_1 \nonumber\\
                 &\leq 1+\frac{2\gamma }{1-\gamma},\nonumber
\end{align*}
where the second inequality follows from that fact that \( \delta_k \) has only one non-zero term. Applying the result to~\eqref{eq:w_k_var}, we have
\begin{align}
    \mathbb{E}[||w_k||^2_2 \mid \mathcal{F}_k ] &\leq 1 +\frac{4 \gamma }{1-\gamma} +  \frac{4\gamma^2}{(1-\gamma)^2} \nonumber \\
                                              &\leq \frac{9}{(1-\gamma)^2} \nonumber,
\end{align}        
which is the desired conclusion.

\import{./appendix}{matrix_inequalities}

\import{./appendix}{app_sec7.tex}



\subsection{Sample Complexity under i.i.d. observation models}

\begin{lemma}[Sample Complexity with respect to expectation]
To achieve \( \mathbb{E}[||Q_k - Q^*||_{\infty}] \leq \epsilon \), we need at most following number of samples: 
\begin{equation}
\mathcal{O}\left(\frac{(|\mathcal{S}||\mathcal{A}|)^{8}}{(1-\gamma)^7d_{\min}^5} \frac{1}{\epsilon^2} \right) \nonumber
\end{equation}
\end{lemma}
\begin{proof}
To achieve \( \mathbb{E}[||Q_k - Q^*||_{\infty}] \leq \epsilon \), one sufficient condition using the bound in~\Cref{thm:iid_convergence_rate} is  
\begin{align}
  \mathbb{E} [  ||Q_k - Q^*||_{\infty}] \leq \frac{1}{\sqrt{k+\xi}} \max \left\{\frac{32(|\mathcal{S}||\mathcal{A}|)^4d_{\max}^{\frac{5}{2}}}{(1-\gamma)^{\frac{5}{2}}d_{\min}^{\frac{5}{2}}}||Q^L_0 -Q^*||_2 ,
    \frac{112 (|\mathcal{S}| |\mathcal{A}|)^{\frac{11}{4}} d_{\max}^{\frac{3}{2}}}{(1-\gamma)^{\frac{7}{2}}d_{\min}^{\frac{5}{2}}} \right\}  &\leq \epsilon ,\nonumber
\end{align}
which leads to
\begin{align}
    k &\geq \frac{1}{\epsilon^2} \max \left\{ \frac{32^2(|\mathcal{S}||\mathcal{A}|)^8}{(1-\gamma)^5d_{\min}^5}||Q^L_0 -Q^*||_2^2, \frac{112^2(|\mathcal{S}||\mathcal{A}|)^{\frac{11}{2}}}{(1-\gamma)^7 d_{\min}^5} \right\} \nonumber.
\end{align}
\end{proof}

Now, the convergence in expectation can lead to the following probability tail bound.
\begin{lemma}[Sample complexity with respect to probability tail bound]\label{prop:sample_complexity_iid}
With probability at least \( 1- \delta\), we can achieve \( ||Q^k-Q^*||_{\infty} \leq \epsilon \)  with at most following number of samples:
\begin{align}
   \mathcal{O}\left( \frac{(|\mathcal{S}||\mathcal{A}|)^{8}}{(1-\gamma)^7d_{\min}^5}\frac{1}{\epsilon^2\delta^2} \right) \   \nonumber.
\end{align}
\end{lemma}
\begin{proof}
For simplicity let \(C_I := \max \left\{\frac{16(|\mathcal{S}||\mathcal{A}|)^4d_{\max}^{\frac{5}{2}}}{(1-\gamma)^{\frac{5}{2}}d_{\min}^{\frac{5}{2}}}||Q^L_0 -Q^*||_2 ,
    \frac{112 (|\mathcal{S}| |\mathcal{A}|)^{\frac{11}{4}} d_{\max}^{\frac{3}{2}}}{(1-\gamma)^{\frac{7}{2}}d_{\min}^{\frac{5}{2}}} \right\} \).Applying Markov inequality to~\Cref{thm:iid_convergence_rate} yields
\begin{align}
 \mathbb{P}(  ||Q_k - Q^*||_{\infty} \geq \epsilon ) &\leq \frac{ \mathbb{E}[||Q_k - Q^*||_{\infty}] }{\epsilon} \nonumber\\
 &\leq \frac{1}{\epsilon}\frac{C_I}{\sqrt{k+\xi}}  \label{ineq:i.i.d_sample_1}
\end{align}
The corresponding complement of the event satisfies
\begin{align*}
    \mathbb{P}(||Q_k-Q^*||_\infty \leq \epsilon) &= 1 - \mathbb{P}(||Q_k-Q^*||_{\infty} \geq \epsilon ) \\
                                                 &\geq 1- \frac{1}{\epsilon}\frac{C_I}{\sqrt{k+\xi}},
\end{align*}
where the last inequality follows from~\eqref{ineq:i.i.d_sample_1}. Now, to achieve \(\mathbb{P} ( || Q_k -Q^*||_{\infty} \leq \epsilon ) \geq 1- \delta  \), we lower bound the above inequality with \(1-\delta\), which leads to
\begin{align*}
  \frac{1}{\epsilon} \frac{C_I}{\sqrt{k+\xi}} &\leq \delta \nonumber.
\end{align*}
The above inequality results in
\begin{align}
  k \geq  \frac{1}{\epsilon^2\delta^2} \max \left\{ \frac{16^2(|\mathcal{S}||\mathcal{A}|)^8}{(1-\gamma)^5d_{\min}^5}||Q^L_0 -Q^*||_2^2, \frac{112^2(|\mathcal{S}||\mathcal{A}|)^{\frac{11}{2}}}{(1-\gamma)^7 d_{\min}^5} \right\} \nonumber.
\end{align}
This completes the proof.
\end{proof}


\subsection{Sample Complexity under Markovian noise}\label{app:sec:sample_complexity}

In this subsection, we derive sample complexity under Markovian noise.
As preliminary results, some lemmas are introduced first. 
\begin{lemma}\label{lem:mixing_time_bound}
The mixing time, $\tau^{\mix}(\alpha_k)$, has the following bound:
\begin{align}
     \tau^{\mix}(\alpha_k) \leq  \frac{1}{\log\rho} \log \frac{m(k+\xi)}{\theta} +1 \nonumber.
\end{align}
\end{lemma}

\begin{proof}
By the definition of the mixing time in~\eqref{def:mixing_time}, \( \tau^{\mix}(\alpha_k ) \) satisfies
 \begin{align}
     m\rho^{\tau^{\mix}(\alpha_k )} \leq \alpha_k . \label{ineq:app:mix}
 \end{align}
 Hence, simply rearranging the terms, we have the following inequalities:
\begin{align}
     \tau^{\mix}(\alpha_k ) \log \rho + \log m &\leq \log \alpha_k \nonumber \\
     \log \frac{1}{\rho} \log \frac{m}{\alpha_k} &\leq  \tau^{\mix}(\alpha_k ) \nonumber\\
     \frac{1}{\log \frac{1}{\rho}} \log \frac{m}{\alpha_k} &= \frac{1}{\log \frac{1}{\rho}} \log \frac{m(k+\xi)}{\theta}  \leq \tau^{\mix}(\alpha_k ), \label{app:tau_mix_ineq_2}
 \end{align}
where the first inequality follows by taking logarithm on both sides of~\eqref{ineq:app:mix}, the second inequality is obtained by re-arranging the first inequality, and The last inequality is due to the definition of step-size \(\alpha_k\) in~\eqref{eq:step_size}.
From the definition of mixing time in~\eqref{def:mixing_time} , \( \tau^{\mix}(\alpha_k)\) is the minimum integer achieving the bound~\eqref{app:tau_mix_ineq_2}. Therefore we have
 \( \tau^{\mix}(\alpha_k) \leq \frac{1}{\log\rho} \log \frac{m(k+\xi)}{\theta} + 1 \), which concludes the proof. 
\end{proof}

Without loss of generality, to keep simplicity, we assume
\begin{align}
         \tau^{\mix}(\alpha_k) \leq  \frac{1}{\log\rho} \log \frac{m(k+\xi)}{\theta} . \nonumber
\end{align}

\begin{lemma}\label{lem:sample_complexity_expectation_markovian}
To achieve \( \mathbb{E}[||Q^k-Q^*||_{\infty}] \leq \epsilon \), we need at most following number of samples:
\begin{align}
  \tilde{\mathcal{O}}\left( \frac{|\mathcal{S}||\mathcal{A}|)^{\frac{33}{2}} }{(1-\gamma)^{16} d_{\min}^{12}}\frac{1}{\epsilon^2} \right) \nonumber.
\end{align}
where \(\tilde{\mathcal{O}}(\cdot)\) hides logarithmic factors.
\end{lemma}
\begin{proof}
Using the bound in~\Cref{thm:markovian_convergence_rate}, we need to achieve
\begin{align*}
    \mathbb{E}[||Q_k-Q^* ||_{\infty}] &\leq \frac{C^*}{\sqrt{k+\xi}}+\frac{C(|\mathcal{S}||\mathcal{A}|)^{\frac{33}{8}}d_{\max}}{(1-\gamma)^{4} d_{\min}^{3}} \frac{\sqrt{\tau^{\mix}(\alpha_k)}}{\sqrt{k+\xi}} \leq \epsilon.
\end{align*}

A sufficient condition to achieve the bound is to make each term in~\Cref{thm:markovian_convergence_rate} smaller than \( \frac{\epsilon}{2}\). Dosing so leads to the following two inequalities:
    \begin{align}
\frac{C^*}{\sqrt{k+\xi}}   &\leq \frac{\epsilon}{2} ,\nonumber \\
\frac{C(|\mathcal{S}||\mathcal{A}|)^{\frac{33}{8}}d_{\max}}{(1-\gamma)^{4} d_{\min}^{3}} \frac{\sqrt{\tau^{\mix}(\alpha_k)}}{\sqrt{k+\xi}}  &\leq \frac{\epsilon}{2} \nonumber.
    \end{align}
The first inequality leads to
\begin{align}
      k \geq \frac{C}{\epsilon^2}  \max \left( \frac{(|\mathcal{S}||\mathcal{A}|)^{8}}{(1-\gamma)^5d_{\min}^5}||Q^L_0 -Q^*||_2^2 ,
\frac{C (|\mathcal{S}| |\mathcal{A}|)^{\frac{15}{2}}}{(1-\gamma)^{8}d_{\min}^{6}} 
\right)  \label{ineq:markov_sample_copmlexity_1}.
\end{align}
\par
For the second inequality, we first bound the mixing time with a logarithmic term using~\Cref{lem:mixing_time_bound}, and use the fact that \(\log x \leq 2\sqrt{x}\). The second inequality can be achieved by bounding the upper bound
\begin{align*}
      \frac{C(|\mathcal{S}||\mathcal{A}|)^{\frac{33}{8}}}{(1-\gamma)^{4}d_{\min}^3}  \frac{  \sqrt{\tau^{\mix}(\alpha_k)}}{\sqrt{k+\xi}}  &\leq  \frac{C(|\mathcal{S}||\mathcal{A}|)^{\frac{33}{8}}}{(1-\gamma)^{4}d_{\min}^3} \sqrt{\frac{\log \frac{m}{\theta}}{\log\frac{1}{\rho}}}\frac{1}{\sqrt{k+\xi}} \\
      &+  \frac{2C(|\mathcal{S}||\mathcal{A}|)^{\frac{33}{8}}}{(1-\gamma)^{4}d_{\min}^3}\frac{1}{ \sqrt{\log\frac{1}{\rho}}}\frac{\sqrt{2}(k+\xi)^{1/4}}{\sqrt{k+\xi}}  \nonumber.
\end{align*}
Next, bounding each term on the right side of the above inequality with \(\frac{\epsilon}{4}\) leads to
\begin{align}
    k \geq \frac{\log \frac{m}{\theta}}{\log \frac{1}{\rho}} \frac{C(|\mathcal{S}||\mathcal{A}|)^{\frac{33}{4}}}{(1-\gamma)^8d_{\min}^6\epsilon^2} \label{ineq:markov_sample_copmlexity_2},
\end{align}
and
\begin{align}
    k \geq \frac{C d_{\max}^4 (|\mathcal{S}||\mathcal{A}|)^{\frac{33}{2}}}{(1-\gamma)^{16}d_{\min}^{12}}\frac{1}{(\log\frac{1}{\rho})^2}. \label{ineq:markov_sample_copmlexity_3}
\end{align}
Collecting the largest coefficients for \(\frac{1}{1-\gamma}\), \(|\mathcal{S}||\mathcal{A}|\), and \(d_{\min}\) in~\eqref{ineq:markov_sample_copmlexity_1},~\eqref{ineq:markov_sample_copmlexity_2}, and~\eqref{ineq:markov_sample_copmlexity_3}, we get
\begin{align}
  \tilde{\mathcal{O}}\left( \frac{|\mathcal{S}||\mathcal{A}|)^{\frac{33}{2}} }{(1-\gamma)^{16} d_{\min}^{12}}\frac{1}{\epsilon^2} \right) \nonumber.
\end{align}
This completes the proof.
\end{proof}

Now, we derive sample complexity in terms of probability tail bounds using Markov inequality.
The proof is similar to the i.i.d. case, and hence, it is omitted here.

\begin{lemma}\label{lem:markovian_sample_complexity_tail_bound}
With probability at least \( 1- \delta\), we can achieve \( ||Q^k-Q^*||_{\infty} \leq \epsilon \)  with at most following number of samples:
\setlength{\parskip}{0pt}
\setlength{\abovedisplayskip}{0pt}
\setlength{\belowdisplayskip}{0pt}
\setlength{\abovedisplayshortskip}{0pt}
\setlength{\belowdisplayshortskip}{0pt}
\vspace{1em}
\begin{align}
  \mathcal{O}\left(\frac{|\mathcal{S}||\mathcal{A}|)^{\frac{33}{2}} }{(1-\gamma)^{16} d_{\min}^{12}}\frac{1}{\epsilon^2\delta^2} \right) \nonumber.
\end{align}
\end{lemma}
\par

%% file: appendix/matrix_inequalities.tex
\subsection{Proof of~\Cref{lem:G_upper_bound}}\label{app:lem:G_upper_bound}
Before proceeding on, we introduce the following lemma which is required for the main proof. 
\begin{lemma}[\Cite{cormen2022introduction}, Chapter C.1]\label{lem:binom_coef}
For \( 1 \leq k \leq n\), we have
\begin{align*}
    \binom{n}{k}\leq \left( \frac{en}{k} \right)^k ,
\end{align*}
where \( \binom{n}{k} := \frac{n!}{k!(n-k)!}\).
\end{lemma}

Now, we prove the first item in~\Cref{lem:G_upper_bound}. For any $s>0$, we can bound the matrix exponential term in~\eqref{eq:G_solution} as follows:
\begin{align}
   \left\lVert e^{T_{Q^*}s}\right\rVert_{\infty} &= \left\lVert I + T_{Q^*}s + \frac{(T_{Q^*}s)^2}{2!}+ \frac{(T_{Q^*}s)^3}{3!}+ \dots \right\rVert_{\infty} \nonumber \\
   &= \max_{i\in\{1,2,\dots,|\mathcal{S}||\mathcal{A}|\}} \left(\middle|1 + \sum\limits_{j=1}^{|\mathcal{S}||\mathcal{A}|} [T_{Q^*}]_{ij} s + \mathcal{O}(s^2) \middle|\right) \nonumber\\
   &\leq \max_{i\in\{1,2,\dots,|\mathcal{S}||\mathcal{A}|\}} \left( \left| 1 + \sum\limits^{  |\mathcal{S}||\mathcal{A}|}_{j=1} [T_{Q^*}]_{ij} s \right| \right) +\mathcal{O}(s^2) \label{ineq:taylor},
\end{align}
where the first equality expands matrix exponential in terms of matrix polynomials, and the second equality follows from definition of infinity norm. Bounding the first term of~\eqref{ineq:taylor} with the diagonally dominant property in~\Cref{lem:T_q_hurwitz}, we have
\begin{align}\label{ineq:e_taylor_bound}
     \left\lVert e^{T_{Q^*}s}\right\rVert_{\infty}\leq 1- (1 -\gamma) d_{\min} s + \mathcal{O}(s^2) .
\end{align}
Using the last inequality, we can obtain the following relation: 
\begin{align}
    \left\lVert e^{T_{Q^*}s}\right\rVert_{\infty} & = \left\lVert(e^{T_{Q^*}s\frac{1}{n}})^n\right\rVert_{\infty} \nonumber \\
                              &\leq  \left\lVert e^{T_{Q^*}s \frac{1}{n} }\right\rVert_{\infty}^n \nonumber \\
                              &\le \left\{ 1 - (1 - \gamma )d_{\min } s \frac{1}{n} + \mathcal{O}\left( \frac{s^2}{n^2}  \right) \right\}^n \label{ienq:taylor_1},
\end{align}
where the first inequality follows from the sub-multiplicativity of matrix norm, and the second inequality follows from directly applying~\eqref{ineq:e_taylor_bound}. 


Taking the limit on both sides and using the relation, \( \lim_{n \to \infty} ( 1- \frac{a}{n})^n = e^{-a}\) for any $a>0$, we can derive the following bound:
\begin{align*}
\left\| e^{T_{Q^* } s} \right\|_\infty \le& \lim_{n \to \infty } \left\{ 1 - (1 - \gamma)d_{\min} s\frac{1}{n} + \mathcal{O}\left( \frac{s^2}{n^2} \right) \right\}^n \\
=& \lim_{n\to\infty} \sum\limits_{k=0}^n \binom{n}{k} \left(1 - (1 - \gamma)d_{\min} s\frac{1}{n}\right)^{n-k} \left( \mathcal{O}\left( \frac{s^2}{n^2} \right) \right)^{k} \\
\leq &\lim_{n\to\infty} \sum\limits_{k=0}^n (en)^k \left(1 - (1 - \gamma)d_{\min} s\frac{1}{n}\right)^{n-k} \left( \mathcal{O}\left( \frac{s^2}{n^2} \right) \right)^{k}\\
= & \lim_{n\to\infty} \sum\limits_{k=0}^n \left(1 - (1 - \gamma)d_{\min} s\frac{1}{n}\right)^{n-k} \left( \mathcal{O}\left( \frac{s^2}{n} \right)\right)^{k}\\
=&\lim_{n\to\infty} \left ( 1- (1-\gamma)d_{\min}s\frac{1}{n}\right)^n \\
&+\lim_{n\to\infty} \sum\limits_{k=1}^n  \left(1 - (1 - \gamma)d_{\min} s\frac{1}{n}\right)^{n-k}\left( \mathcal{O}\left( \frac{s^2}{n} \right)\right)^{k}\\
\leq & \lim_{n\to\infty} \left ( 1- (1-\gamma)d_{\min}s\frac{1}{n}\right)^n +\lim_{n\to\infty} \sum\limits_{k=1}^n  \left( \mathcal{O}\left( \frac{s^2}{n} \right)\right)^{k}\\
=& \lim_{n\to\infty} \left ( 1- (1-\gamma)d_{\min}s\frac{1}{n}\right)^n +\lim_{n\to\infty} \mathcal{O}\left(\frac{s^2}{n}\right) \nonumber \\
 =& e^{-(1 - \gamma)d_{\min} s} ,
\end{align*}
where \( \binom{n}{k} := \frac{n!}{k!(n-k)!}\). The first equality follows from the binomial theorem, and the second inequality follows from~\Cref{lem:binom_coef}, which bounds the binomial coefficient. The second last equality follows from summation of geometric series.
Therefore, we have
\begin{align}
    \left\lVert e^{T_{Q^*}s}\right\rVert_{\infty}\leq e^{-(1-\gamma)d_{\min}s} . \label{ineq:||e||_bound}
\end{align}
Next, we bound the spectral norm of \(G\) as follows:
\begin{align}
  \left\lVert G \right\rVert_2 &=  \int_0^\infty  \left\lVert(e^{T_{Q^*} t} )\right\rVert^2_2 dt \nonumber\\
            &\leq |\mathcal{S}||\mathcal{A}|\int_0^\infty \left\lVert(e^{T_{Q^*} t} )\right\rVert^2_{\infty} dt \nonumber \\
            &\leq\int_0^\infty e^{-2(1-\gamma)d_{\min}t} dt  \nonumber \\
            &\leq \frac{|\mathcal{S}||\mathcal{A}|}{2(1-\gamma)d_{\min}} \nonumber,
\end{align}
where the first equality comes from the fact that \( \left\lVert e^{(T_{Q^*t})^{\top}} \right \rVert_2 = \left\lVert e^{T_{Q^*t}} \right\rVert_2 \), and the second inequality follows from~\eqref{ineq:||e||_bound}. The lower bound of \(||G||_2 \) follows from applying simple triangle inequality to~\eqref{eq:cale}.
\begin{align*}
    ||G||_2 \geq   \frac{1}{2||T_{Q^*}||_2}\geq  \frac{1}{2\sqrt{|\mathcal{S}|\mathcal{A}|}||T_{Q^*}||_{\infty}}\geq  \frac{1}{4\sqrt{|\mathcal{S}|\mathcal{A}|}d_{\max}},
\end{align*}
where the last inequality follows from~\Cref{lem:T_q_hurwitz}. Now, we prove the second item, bound on \( ||G^{-1}||_2\). Multiplying \(G^{-1}\) on the right and left of~\eqref{eq:cale}, we have
\begin{align}
    T_{Q^*}G^{-1}+ G^{-1} T^{\top}_{Q^*} = -  G^{-2}. \label{eq:inv_cale}
\end{align}
Now taking the spectral norm on~\eqref{eq:inv_cale}, and using the triangle inequality, we have
\begin{align}
     ||G^{-2}||_2 \leq 2||T_{Q^*}||_2 ||G^{-1}||_2 . \label{ineq:G^-1_1}
\end{align}
Since \( G^{-1}\) is symmetric positive definite, it follows that \( ||G^{-2}||_2 = ||G^{-1}||_2^2 \).
Therefore, we have
\begin{align}
     ||G^{-1}||_2  &\leq 2(|\mathcal{S}||\mathcal{A}|)^{\frac{1}{2}}||T_{Q^*}||_{\infty}  \nonumber \\
                   &\leq 4(|\mathcal{S}||\mathcal{A}|)^{\frac{1}{2}}d_{\max} \nonumber, 
\end{align}
where the first inequality follows from~\eqref{ineq:G^-1_1} and the fact that \(||G^{-1}||_2>0\), and the last inequality comes from~\Cref{lem:T_q_hurwitz}. Next, the lower bound of \( ||G^{-1}||_2 \) can be obtained as follows:
\begin{align}
   ||G^{-1}||_2 = \frac{1}{\sigma_{\min}(G)} \geq \frac{2(1-\gamma)d_{\min}}{|\mathcal{S}||\mathcal{A}|} \nonumber,
\end{align}
where the first equality is due to the fact that \(G\) is positive symmetric definite, and the last inequality follows from the upper bound on \(||G||_2\). This completes the proof.

\subsection{Proof of~\Cref{lem:B_upper_bound}}\label{app:lem:B_upper_bound}

The upper bound of \(||B||_2\) follows from the following simple algebraic inequalities:
\begin{align*}
    ||B||_2 &\leq ||G^{1/2}||_2 ||T_{Q^*}||_2 ||G^{-1/2}||_2\\
    &=  ||G||_2^{1/2}||T_{Q^*}||_2 ||G^{-1}||_2^{1/2}\\
    &\leq  \sqrt{|\mathcal{S}||\mathcal{A}|}||G||_2^{1/2} ||T_{Q^*}||_{\infty} ||G^{-1}||_2^{1/2}\\
    &\leq   \sqrt{|\mathcal{S}||\mathcal{A}|} \left(\frac{|\mathcal{S}||\mathcal{A}|}{2(1-\gamma)d_{\min}}  \right)^{\frac{1}{2}}
  2d_{\max}
    \left(2(|\mathcal{S}||\mathcal{A}|)^{\frac{1}{2}}{(1-\gamma)d_{\min}}\right)^{1/2}\\
    &= 2(|\mathcal{S}||\mathcal{A}|)^{\frac{5}{4}} d_{\max} ,
\end{align*}
where the first inequality follows from the sub-multiplicativity of the norm, and the fact that \(G^{-1}\) is positive symmetric definite. In addition, the second inequality follows from the upper bound on \(||G||_2\) in~\Cref{lem:G_upper_bound} and \(||T_{Q^*}||_{\infty}\) in~\Cref{lem:T_q_hurwitz}. 

The lower bound can be derived from applying triangle inequality to~\eqref{eq:lyapunov_eq_B} as follows:
\begin{align}
    ||B||_2 &\geq  \frac{1}{2} ||G^{-1}||_2  \geq \frac{(1-\gamma)d_{\min}}{|\mathcal
    S||\mathcal{A}|} \nonumber,
\end{align} 
where the last inequality follows from~\Cref{lem:inv_G_bound}. This completes the proof.

%% file: appendix/app_sec7.tex

\subsection{Proof of~\Cref{lem:markovian_noise_bound}}\label{app:lem:markovian_noise_bound}
The bound on \(||w_k||_{\infty}\) can be proved with the same logic as in~\Cref{lem:w_k_bound}. For \(\mathbb{E}[||w_k ||_2^2\mid \mathcal{F}_k]\), we have
\begin{align*}
    \mathbb{E}[||w_k||^2_2\mid\mathcal{F}_k] &\leq |\mathcal{S}||\mathcal{A}|\mathbb{E}[||w_k||_{\infty}^2|\mathcal{F}_k] \\
    &\leq \frac{16|\mathcal{S}||\mathcal{A}|}{(1-\gamma)^2},
\end{align*}
where \( \mathcal{F}_k\) is a \(\sigma\)-field induced by \(\{ (s_i,a_i,s_{i+1},r_i, Q_i)\}_{i=0}^k\), the first inequality follows from \(||\cdot||_2 \leq (|\mathcal{S}||\mathcal{A}|)^{\frac{1}{2}}||\cdot||_{\infty}\), and the last inequality is due to the bound on \(||w_k||_{\infty}\). This completes the proof.

\subsection{Proof of~\Cref{lem:bdd_xi}}\label{app:lem:bdd_xi}
Before proceeding, an important lemma is fist established. 
\begin{lemma}\label{lem:w_q_lipschitz}
\(w_k(Q)\) is Lipschitz with respect to \(Q\) :
\begin{align}
    ||w_k(Q)- w_k (Q') ||_{\infty} \leq 4 ||Q-Q'||_{\infty}  \nonumber
\end{align}
\end{lemma}
\begin{proof}
The proof follows from simple algebraic inequalities:
    \begin{align*}
         ||w_k(Q)- w_k (Q') ||_{\infty} &= ||(\gamma (e_{a_k} \otimes e_{s_k})(e_{s_{k+1}})^{\top}-\gamma D_{\infty}P)(\Pi_{Q}Q -\Pi_{Q'}Q')||_{\infty} \\
         &+||(e_{a_k} \otimes e_{s_k} ) (e_{a_k} \otimes e_{s_k})^{\top} -D_{\infty}) (Q-Q') ||_{\infty} \nonumber\\
         &\leq (1+\gamma) (1+ d_{\min}) ||Q-Q'||_{\infty} \nonumber\\
         &\leq 4||Q-Q'||_{\infty}.
    \end{align*}
This completes the proof.
\end{proof}

Now we prove the boundedness of \( \xi_k(z_k,Q_k)\). The proof simply uses boundedness of \(z_k\) as follows:
    \begin{align*}
    |\xi_k(z_k,Q_k)| &\leq ||X_k z_k||_2 ||G^{1/2}w_k||_2 \nonumber \\
    &\leq  ||z_k ||_{2}||G^{1/2}w_k||_2  \nonumber \\
    &\leq \frac{8(|\mathcal{S}||\mathcal{A}|)^{\frac{3}{2}}}{(1-\gamma)^{\frac{5}{2}}d_{\min}^{\frac{3}{2}}} \frac{(|\mathcal{S}||\mathcal{A}|)^{\frac{1}{2}}}{(1-\gamma)^{\frac{1}{2}}d_{\min}^{\frac{1}{2}}} \frac{4|\mathcal{S}||\mathcal{A}|}{(1-\gamma)} \nonumber \\
    &= \frac{32 (||\mathcal{S}||\mathcal{A}|)^3}{(1-\gamma)^4d_{\min}^2 },
    \end{align*}
where the first inequality follows from H\"older's inequality, the second inequality is due to the fact that \(||X_k||_2 \leq 1\), and the last inequality follows from combining~\Cref{lem:bound_z_k},~\Cref{lem:markovian_noise_bound}, and~\Cref{lem:bound on singular value of X_k}.

Moreover, the Lipchitzness of \(\xi_k(z,Q)\) can be also proved using boundedness of \(z_k\) and \(w_k\) as follows:
\begin{align}
           |\xi_k (z,Q) - \xi_k (z',Q')| &=  | z^{\top}X_k^{\top}G^{1/2}w_k(Q) -z'^{\top}X_k^{\top}G^{1/2}w_k(Q') | \nonumber \\
           &\leq |z^{\top}X^{\top}_kG^{1/2} ( w_k (Q) - w_k (Q') )| + |w_k(Q')^{\top}G^{1/2}(X_k z - X_k z') | \nonumber\\
           &\leq ||G^{1/2}X_k z||_2 ||w_k (Q) - w_k (Q')||_2 +
           ||X^{\top}_kG^{1/2}w_k(Q') ||_2 ||z-z'||_2 \nonumber\\
           &\leq 4|\mathcal{S}||\mathcal{A}|||G^{1/2}X_k z||_2||Q-Q'||_{\infty} + ||X^{\top}_kG^{1/2}w_k(Q') ||_2 ||z-z'||_2 \nonumber\\
           &\leq \frac{16(|\mathcal{S}||\mathcal{A}|)^3}{(1-\gamma)^3d_{\min}^2} ||Q-Q'||_{\infty} + \frac{4(|\mathcal{S}||\mathcal{A}|)^{\frac{3}{2}}}{(1-\gamma)^{\frac{3}{2}}d_{\min}^{\frac{1}{2}}}||z-z'||_{\infty} \nonumber,
\end{align}
where the first inequality is due to the fact that \(a_1^{\top}b_1 - a_2^{\top}b_2 = (a_1-a_2)^{\top}b_1 + a_2^{\top}(b_1-b_2) \). The second inequality follows from H\"older's inequality, the third inequality is due to the Lipchitzness of \(w_k(Q)\) in~\Cref{lem:w_q_lipschitz}, and the last inequality comes from collecting bounds on each terms. This completes the proof.
